\documentclass[lettersize,journal]{IEEEtran}
\usepackage{amsmath,amsfonts}

\usepackage{algorithm}
\usepackage{algorithmic}
\usepackage{array}
\usepackage[caption=false,font=normalsize,labelfont=sf,textfont=sf]{subfig}
\usepackage{textcomp}
\usepackage{stfloats}
\usepackage{hyperref}
\usepackage{xcolor}
\usepackage{graphicx}
\usepackage{relsize}
\newcommand{\pub}[1]{%
    \textcolor{gray}{\scalebox{0.8}{[#1]}}%
}
\usepackage{url}
\usepackage{verbatim}
\usepackage{graphicx}
\usepackage{cite}
\usepackage{bbding}
\usepackage{pifont}
\usepackage[table]{xcolor} 
\usepackage{makecell}
\definecolor{myyellow}{HTML}{ffffb2} 
\definecolor{myorange}{HTML}{ffe4cf} 
\definecolor{myred}{HTML}{ffb2b2} 
\definecolor{mygreen}{HTML}{00ff00} 
\newcommand{\greencheck}{\textcolor{green}{\ding{51}}}
\newcommand{\redcross}{\textcolor{red}{\ding{55}}}
\usepackage{multirow} 
\usepackage[T1]{fontenc} 
\usepackage{orcidlink}

\begin{document}

\title{HBSplat: Robust Sparse-View Gaussian Reconstruction with Hybrid-Loss Guided Depth and Bidirectional Warping}

\author{
Yu~Ma~\orcidlink{0009-0005-7650-3266}, 
Guoliang~Wei~\orcidlink{0000-0003-2928-4142}, 
Haihong~Xiao~\orcidlink{0000-0002-3543-9262},
Yue~Cheng~\orcidlink{0009-0005-6471-6520},


\thanks{Yu Ma, Guoliang Wei and Yue Cheng are with the University of Shanghai for Science and Technology, Shanghai 200093, China, (e-mail: yummychina@163.com; guoliang.wei@usst.edu.cn; chengyue916@163.com).}
\thanks{Haihong Xiao is with the Hefei University of Technology, Hefei 230009, China, (e-mail: hhxiao@hfut.edu.cn).}

}




\maketitle
\begin{abstract}
Novel View Synthesis (NVS) from sparse views presents a formidable challenge in 3D reconstruction, where limited multi-view constraints lead to severe overfitting, geometric distortion, and fragmented scenes. While 3D Gaussian Splatting (3DGS) delivers real-time, high-fidelity rendering, its performance drastically deteriorates under sparse inputs, plagued by floating artifacts and structural failures. To address these challenges, we introduce HBSplat, a unified framework that elevates 3DGS by seamlessly integrating robust structural cues, virtual view constraints, and occluded region completion. Our core contributions are threefold: a Hybrid-Loss Depth Estimation module that ensures multi-view consistency by leveraging dense matching priors and integrating reprojection, point propagation, and smoothness constraints; a Bidirectional Warping Virtual View Synthesis method that enforces substantially stronger constraints by creating high-fidelity virtual views through bidirectional depth-image warping and multi-view fusion; and an Occlusion-Aware Reconstruction component that recovers occluded areas using a depth-difference mask and a learning-based inpainting model. Extensive evaluations on LLFF, Blender, and DTU benchmarks validate that HBSplat sets a new state-of-the-art, achieving up to 21.13 dB PSNR and 0.189 LPIPS, while maintaining real-time inference. Code is available at: \url{https://github.com/eternalland/HBSplat}.

\end{abstract}

\begin{IEEEkeywords}
Novel View Synthesis, Sparse Views, 3D Gaussian Splatting, Depth Estimation, Bidirectional Warping
\end{IEEEkeywords}

\begin{figure}
\centering
\includegraphics[width=3.4in]{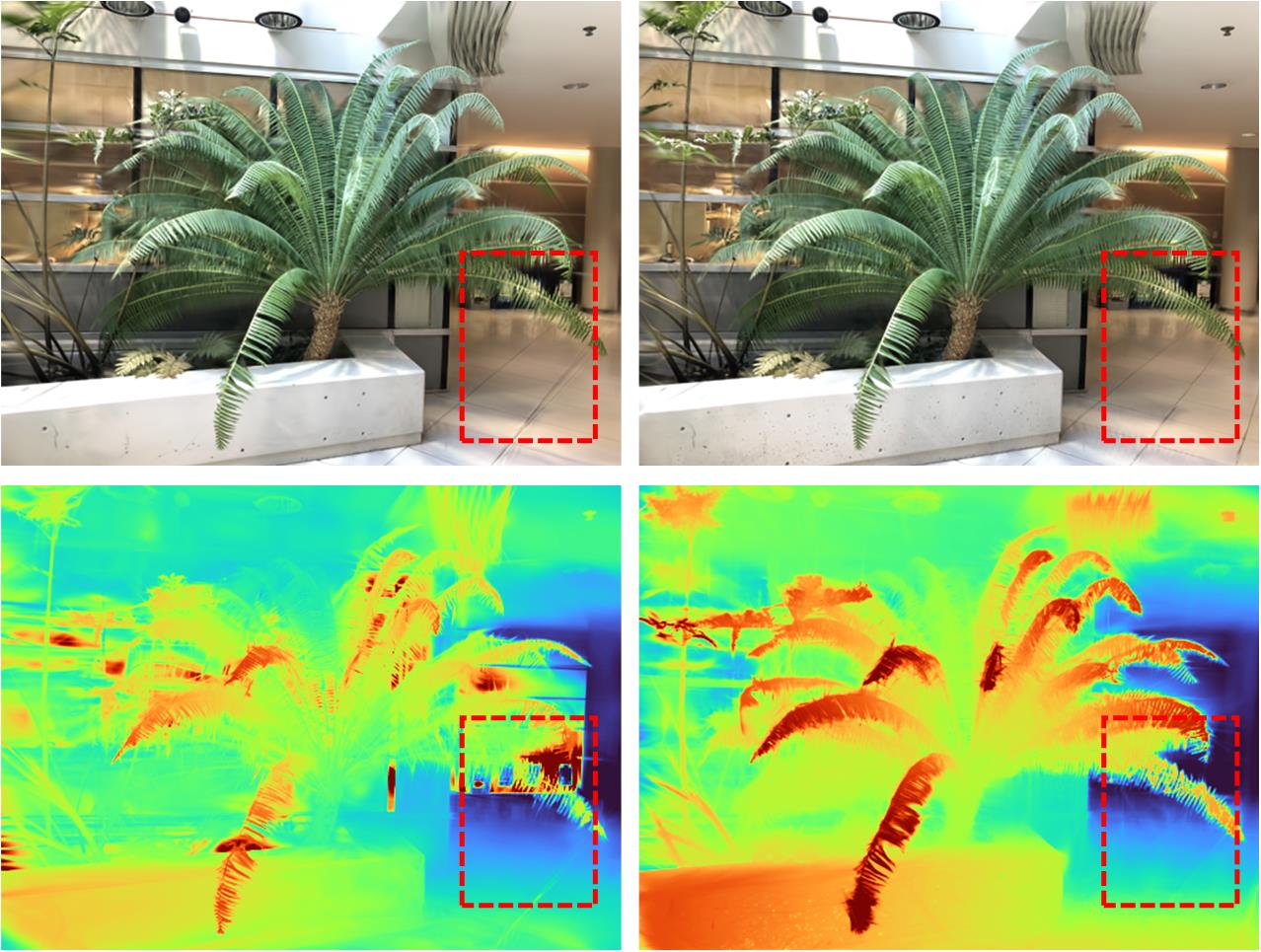}

\begin{center}
\footnotesize  
 (a) Under Sparse-View Reconstruction: Rendered images and depth maps comparison between MCGS (left) and HBSplat (right).
\end{center}
\includegraphics[width=3.4in]{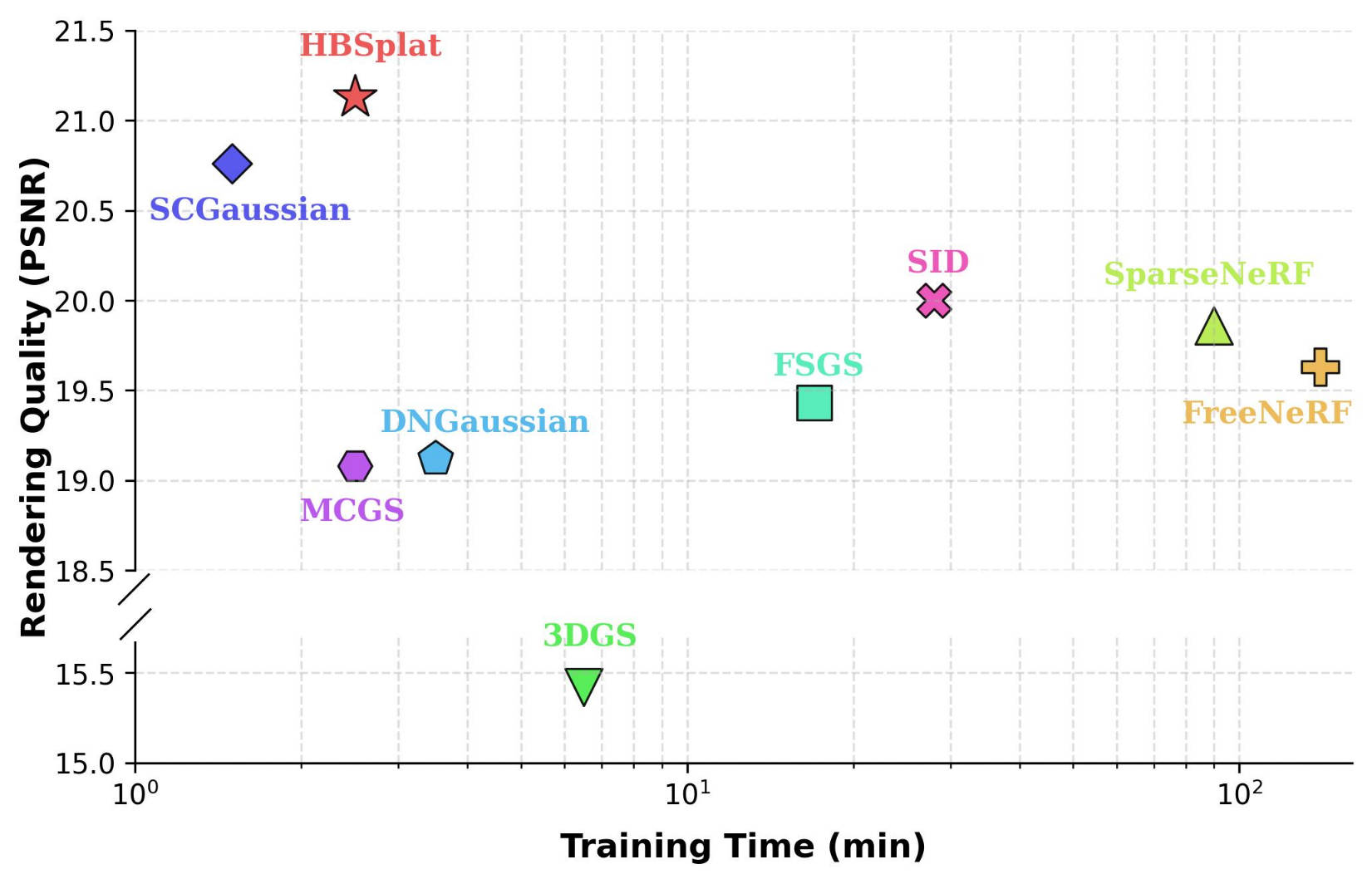}
\begin{center}
\vspace{-3mm}
\footnotesize  
(b) Training Efficiency vs. Rendering Quality.
\end{center}
\vspace{-3mm}
\caption{(a) MCGS outputs (left) vs. HBSplat outputs (right) for both rendered image and depth map. (b) presents an efficiency-quality scatter plot comparing HBSplat with various baseline methods.}
\vspace{-3mm}
\label{page1-f}
\end{figure}

\section{Introduction}
\IEEEPARstart{N}{ovel} View Synthesis (NVS) aims to generate realistic 3D scenes from 2D images, serving as a cornerstone for applications in virtual reality\cite{ref_new1}, robotics\cite{ref_new2}, and autonomous driving\cite{ref_new3}. While Neural Radiance Fields (NeRF)~\cite{ref1} achieve photorealistic quality through implicit neural modeling, they require dozens of input views and suffer from high computational costs, hindering real-time use. 3D Gaussian Splatting (3DGS)~\cite{ref2} emerged as an efficient explicit alternative, leveraging optimizable Gaussians and differentiable rasterization for high-fidelity, real-time rendering. However, under sparse-view conditions (e.g., 3--5 images), insufficient multi-view constraints cause 3DGS to overfit, leading to geometric distortions, floating artifacts, background collapse, and incomplete occlusion handling, especially in complex 360° scenes.

NVS under sparse-view conditions primarily grapples with three fundamental challenges: (1) Inadequate Geometry Initialization: Vanilla 3DGS relies on COLMAP-generated sparse point clouds for initializing Gaussian primitives. Under sparse inputs, these points become insufficient and noisy, failing to support meaningful Gaussian reconstruction. This often causes primitives to over-scale in an attempt to cover surrounding regions, rather than densify appropriately through cloning or splitting under sparse conditions, resulting in blurry and unstable geometry. (2) Under-Constrained Optimization: Limited training views provide weak supervisory signals, leading to overfitting and poor generalization. The model often produces floaters and inconsistent geometry when rendered from novel poses. (3) Occlusion and Unseen Regions: Areas unobserved in all training views cannot be recovered without strong priors or explicit reasoning, resulting in holes or incorrect content in disoccluded regions.

To address the challenges of sparse-view NVS, recent studies have proposed various approaches\cite{ref7,ref9,ref11,ref_new5,ref21, ref35, ref_new4}. NeRF-based methods mitigate overfitting through depth regularization, semantic consistency, and pre-trained priors, but they remain constrained by long training times and slow inference. 3DGS-based methods exploit the flexibility of explicit representations, exploring depth-guided optimization, multi-view consistency constraints, and diffusion model priors, significantly improving efficiency and quality. For instance, depth regularization methods\cite{ref35} provide geometric priors via monocular depth estimation or Multi-View Stereo (MVS) with virtual view generation and co-regularization enhance multi-view consistency. Nevertheless, existing methods still face limitations in geometric consistency, occlusion handling, and detail recovery under sparse views, particularly in complex scenes, where scale inconsistencies in monocular depth priors and low overlap in MVS can lead to inaccurate geometry.

To tackle these challenges, we propose HBSplat, a novel 3DGS-based framework designed for high-quality and efficient novel view synthesis from sparse input views. The framework introduces three key innovations: First, a Hybrid-Loss Depth Estimation module utilizes a dense matcher to extract sufficient initialization points, followed by a novel composite loss integrating reprojection, point propagation, and Total Variation (TV) smoothness constraints to obtain robust depth estimates, significantly enhancing multi-view consistency. The point propagation constraint specifically strengthens geometric consistency by exploiting indirect correspondences across multiple views (e.g., deriving constraints between points $a$ and $c$ via their common matches to point $b$ in overlapping images). Second, a Bidirectional Warping Virtual View Synthesis strategy is employed to generate photometrically and geometrically consistent virtual views, providing additional supervision signals to improve robustness. Third, an Occlusion-Aware Reconstruction component addresses occlusion challenges through depth-difference-based foreground mask and a dedicated inpainting model, effectively reconstructing missing regions by prioritizing background content. Figure \ref{page1-f} presents a comparison between HBSplat and other baseline methods.

The main contributions of this work are the following:
\begin{itemize}
\item We introduce a comprehensive framework that seamlessly integrates depth estimation, virtual view synthesis, and occlusion-aware processing into 3DGS to achieve high-quality NVS from sparse inputs.
\item We develop a robust Hybrid-Loss Depth Estimation module that leverages dense matching priors through a composite loss function, enforcing multi-view geometric consistency by combining reprojection, point propagation, and Total Variation constraints.
\item We propose a Bidirectional Warping strategy to generate photometrically and geometrically consistent virtual training views, thereby significantly enhancing reconstruction quality.
\item We design an innovative Occlusion-Aware Reconstruction mechanism that addresses severe occlusions through a depth-difference-based foreground mask and a learning-based inpainting model to effectively recover background content in heavily occluded regions.
\end{itemize}

Extensive experiments on LLFF, IBRNet, Blender, DTU, and Tanks\&Temples show that HBSplat achieves new state-of-the-art performance across metrics including PSNR (up to 21.13dB) and LPIPS (as low as 0.189), while maintaining real-time rendering speeds. The framework demonstrates remarkable robustness across a spectrum of settings, from forward-facing to 360° scenes.

\begin{figure*}[!t]
\centering
\includegraphics[width=6.7in]{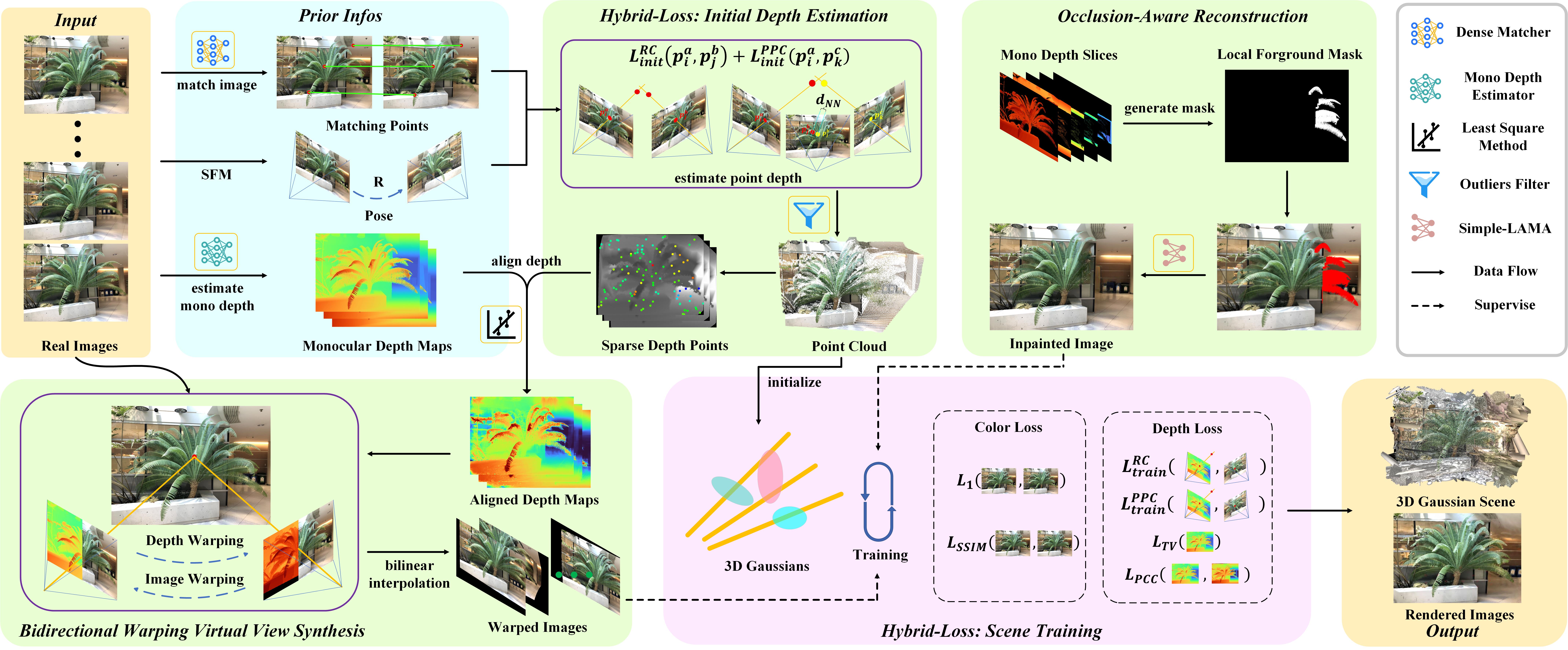}

\caption{HBSplat pipeline. First, sparse input images are processed by dense matching, structure from motion (SfM), and monocular depth estimation to obtain correspondences, camera poses, and depth maps. The Hybrid-Loss module fuses these inputs to produce robust point-wise depths. Subsequent least-squares optimization aligns the point cloud with monocular depths, recovering the metric scale. The Bidirectional Warping module leverages these aligned depths and images to synthesize novel virtual training views through depth-image warping and interpolation. Simultaneously, the Occlusion-Aware Reconstruction component restores missing background content in occluded regions using learning-based inpainting model (Simple-LAMA) guided by local foreground mask. Finally, the framework reconstructs the 3D Gaussian scene by optimizing a joint loss that combines color and depth supervision.}
\label{pipeline0-f}
\end{figure*}

\section{Related Works}
\subsubsection{NeRF-Based NVS Methods} 
NeRF~\cite{ref1} implicitly represent scenes via Multi-Layer Perceptrons (MLPs), achieving photorealistic NVS through volume rendering and becoming a dominant approach in recent years. Subsequent improvements include Mip-NeRF~\cite{ref3}, which mitigates aliasing via conical frustum rendering; Mip-NeRF360~\cite{ref4}, extending the framework to 360° unbounded scenes; and InstantNGP~\cite{ref5}, which reduces training time to seconds using multi-resolution hash encoding. However, these methods generally require dense view inputs (often dozens of images) and suffer from slow training and inference. Moreover, in sparse-view settings, NeRF-based approaches are prone to overfitting, resulting in blurry or inconsistent novel views.

To address sparse-view challenges, various regularization strategies have been proposed. RegNeRF \cite{ref6} mitigates overfitting with color and depth consistency regularization for unseen views, DietNeRF \cite{ref7} enforces semantic consistency using CLIP \cite{ref8} embeddings, and SparseNeRF \cite{ref9} enhances geometry with scale-invariant losses from monocular depth estimation. ViP-NeRF \cite{ref10} and FreeNeRF \cite{ref11} improve training stability via visibility regularization and frequency control, respectively. ReconFusion \cite{ref12} integrates diffusion models to generate additional views while struggling with view consistency. Pose-free reconstruction methods like iNeRF \cite{ref13}, NeRFmm \cite{ref14}, and BARF \cite{ref15} jointly optimize camera poses and scene representations but are limited under complex camera trajectories (e.g., 360° scenes). These methods rely on volume rendering, resulting in low efficiency unsuitable for real-time applications.

\subsubsection{3DGS-Based NVS Methods} 3DGS represents scenes with explicit Gaussian primitives, achieving real-time rendering via differentiable rasterization, surpassing NeRF in efficiency and quality. 3DGS excels with dense views, supporting tasks like text-to-3D generation \cite{ref16} and dynamic scene modeling \cite{ref18,ref19}. However, in sparse views, 3DGS suffers from insufficient initial point clouds and limited multi-view constraints, leading to artifacts and geometry degradation.

Recent studies optimize sparse-view 3DGS with various strategies. FSGS \cite{ref20} enhances geometry via unpooling and monocular depth priors but suffers from long training times, limiting its applicability in time-sensitive scenarios. GaussianObject \cite{ref21} proposes structure-prior-aided initialization, needing only 4 views, and uses diffusion models to repair occluded regions. DNGaussian \cite{ref22} optimizes Gaussian positions with global-local depth normalization, addressing scale inconsistencies. CoherentGS \cite{ref23} enhances Gaussian consistency using an implicit decoder and smoothness loss, filling occluded regions. CoR-GS \cite{ref24} suppresses Gaussian field disagreements via co-regularization, improving geometric consistency. LM-Gaussian \cite{ref25} integrates stereo priors \cite{ref27} and diffusion models\cite{ref28} for iterative detail refinement. MVPGS \cite{ref29} leverages MVS \cite{ref30} and Vision Transformers (ViTs) \cite{ref31} to excavate multi-view cues, optimizing initialization and appearance. MCGS \cite{ref32} introduces a sparse matcher and progressive pruning for multi-view consistency. SCGaussian \cite{ref36}introduces a hybrid Gaussian representation with structure-consistent optimization and matching priors to enhance 3D consistency. Binocular-Guided 3DGS \cite{ref33} constrains rendered depths with binocular stereo consistency, improving view consistency. SID\cite{ref39}, as an enhanced version of FSGS, incorporating semantic regularization from DINO-ViT\cite{ref26} features and local depth constraints. While existing methods have made progress in specific areas, fundamental challenges including monocular depth scale inconsistencies, robust occlusion handling, and high-fidelity detail recovery in complex scenes have yet to be adequately resolved.

Our HBSplat method presents a unified approach for high-quality sparse-view synthesis, integrating three core technical contributions that advance beyond previous methods. 
The proposed Hybrid-Loss Depth Estimation incorporates multi-view geometric constraints including point propagation, offering stronger robustness than SCGaussian\cite{ref36}, which relies only on reprojection error. The Bidirectional Warping method produces more complete virtual views compared to the forward warping used in MVPGS\cite{ref29}, avoiding hole-filling and post-inpainting. Furthermore, the Occlusion-Aware Reconstruction module performs consistent occlusion handling within a single training stage, unlike the complex two-stage diffusion-based refinement in LM-Gaussian\cite{ref25}.
The unified framework jointly addresses depth, consistency, and occlusion challenges under extreme sparsity, demonstrating clear improvements in rendering quality and operational efficiency.

\section{Methods}
This section details the HBSplat framework, which is structured as follows:
\uppercase\expandafter{\romannumeral3}-A. Preliminaries outlines the 3DGS fundamentals.
\uppercase\expandafter{\romannumeral3}-B.  Dense matching for sufficient initialization, ray-based 3DGS Optimization reduces degrees of freedom to prevent overfitting.
\uppercase\expandafter{\romannumeral3}-C. Hybrid-Loss Depth Estimation improves geometric consistency via hybrid loss fusion.
\uppercase\expandafter{\romannumeral3}-D. Bidirectional Warping generates virtual views to enhance photometric and geometric constraints.
\uppercase\expandafter{\romannumeral3}-E. Occlusion-Aware Reconstruction handles occlusions using depth-difference mask and inpainting model. 
Figure \ref{pipeline0-f} illustrates the pipeline of the proposed HBSplat framework.

\subsection{Preliminary for 3D Gaussian Splatting}
HBSplat leverages 3DGS~\cite{ref2} to represent 3D scenes with explicit Gaussian primitives, enabling real-time NVS through differentiable rasterization. Each Gaussian primitive is defined by a set of differentiable parameters: position vector $\mu$, rotation quaternion $q$, scaling matrix $s$, opacity $\alpha$, and spherical harmonic coefficients $sh$ for view-dependent color, denoted as $\mathcal{G} = \{G_i : \mu_i, q_i, s_i, \alpha_i, sh_i\}_{i=1}^P $. The position and shape of each Gaussian follow a Gaussian distribution, formulated as:
\begin{equation}
G(x) = e^{-\frac{1}{2}(x-\mu)^T\Sigma^{-1}(x-\mu)}.
\end{equation}

To ensure the covariance matrix $\Sigma$ is positive semi-definite, it is decomposed into a scaling matrix $s$ and a rotation matrix $R$ represented by quaternion $q$, reducing optimization complexity:
\begin{equation}
\Sigma = RS S^T R^T.
\end{equation}

For projection into the image plane, the view transformation matrix $W$ and the projective transformation Jacobian $J$ are applied to map the 3D Gaussian as follows:
\begin{equation}
\Sigma' = J W \Sigma W^T J^T.
\end{equation}

Pixel color $C$ is computed via volume rendering, blending $N$ ordered overlapping Gaussians:
\begin{equation}
C = \sum_{i \in N} c_i \alpha_i \prod_{j=1}^{i-1} (1-\alpha_j'),
\end{equation}
where  $c_i$ is the color of the $i$-th Gaussian. $\alpha_j' = \alpha_j G(P)$ is derived from the projected 2D Gaussian $\Sigma'$. Pixel depth $D$ is similarly computed: 
\begin{equation}
D = \sum_{i \in N} d_i \alpha_i \prod_{j=1}^{i-1} (1-\alpha_j'),
\end{equation}
where $d_i$ is the depth of the $i$-th Gaussian.

\subsection{Dense Matching and Ray-Based Gaussian Splatting}
Vanilla 3DGS relies on sparse 3D points from COLMAP, which become critically insufficient under sparse-view conditions, leading to initialization failure and overfitting. Although some methods employ MVSNet~\cite{ref41} or VGGT~\cite{ref42} to generate dense point clouds for initialization, they still suffer from multi-stage cascading errors or scale inaccuracies. Here, we customize two simple yet effective core prior enhancements.

\textbf{Dense Matching for Sufficient Initialization}: The standard sparse feature matching in COLMAP is replaced with a dense matching network. By leveraging inter-image correlations rather than independently extracted feature points, this approach produces denser and more reliable correspondences, significantly improving the quality and coverage of the initial 3D point cloud for Gaussian reconstruction.

\textbf{Ray-Constrained Optimization}: Gaussian centers are constrained to lie along camera rays, reducing 3D position optimization to a one-dimensional depth adjustment. Each point is parameterized as $r = o + z \cdot d$, where $o$ is the optical center, $d$ is the viewing direction, and $z$ is the depth. This constraint enhances optimization stability, suppresses floating artifacts, and improves convergence, particularly under limited views.

\begin{figure}[!t]
\centering
\includegraphics[width=3.4in]{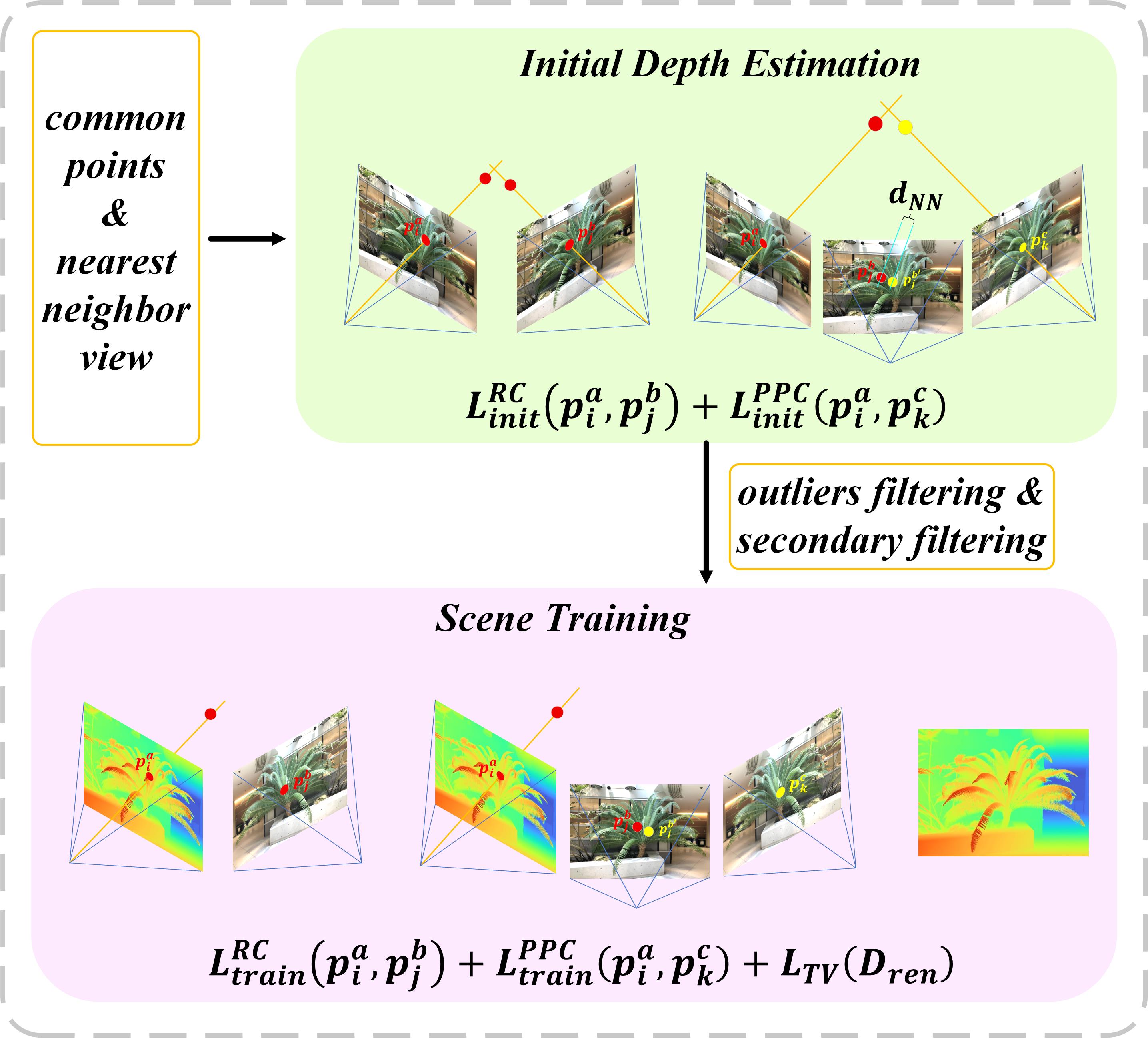}
\caption{Hybrid-Loss Depth Estimation pipeline, which first estimates initial point depth from densely matched points using reprojection and point propagation constraints, then filters outliers. During scene training, rendered depth is refined under reprojection,  point propagation, and TV smoothness constraints. Point propagation constraint are computed from the common points. Nearest-neighbor view and secondary filtering reducing redundant computation.}
\label{pipeline1-f}
\end{figure}

\subsection{Hybrid-Loss Depth Estimation}

Accurate depth estimation is crucial for high-quality scene reconstruction, as it provides the essential geometric information lacking in 2D imagery. However, under sparse-view conditions, the lack of strong multi-view constraints makes high-precision depth estimation difficult through reprojection error minimization alone. To improve depth reliability and multi-view consistency, a hybrid-loss module is proposed that integrates multiple geometric constraints. Figure \ref{pipeline1-f} illustrates the pipeline of the Hybrid-Loss Depth Estimation module.

\subsubsection{Initial Depth Estimation Stage} The depth estimation incorporates two geometric constraints and an outlier filtering mechanism as follows:

\textbf{Reprojection Constraint ($RC$)}: Using dense image matching correspondences, such as point pair $a$-$b$ between view pair $i$-$j$, point $a$ from view $i$ is back-projected into 3D space to obtain point $A$, which is then projected into view $j$ to yield $a'$. The reprojection loss minimizes the Euclidean distance between the reprojected and original points $a'$-$b$:
\begin{equation}
\label{rc1}
L^{RC}_{init} = \sum_{i,j} \| \pi(K, R_{ij}, Ray(p^a_i, z^a_i)) - p^b_j \|_2^2,
\end{equation}
where $p_i$ and $p_j$ are matching image points, $K$ is the camera intrinsic, $R_{ij}$ denotes the relative pose from view $i$ to view $j$, $Ray(\cdot)$ denotes the ray function that generates spatial points, and $\pi(\cdot)$ denotes the projection function.

\textbf{Point Propagation Constraint ($PPC$)}:
Unlike sparse matcher that independently extract feature points, dense matcher generate matching points based on image pairs, making reprojection error constraints alone insufficient for ensuring depth consistency and robustness. To address this, the Point Propagation Constraint is introduced: for matching point pair $a$-$b$ between view pair $i$-$j$, and $b'$-$c$ between view pair $i$-$k$, we use point $b$ and $b'$ in view $j$ as a \textbf{bridge} to establish indirect geometric consistency between $a$ and $c$ across views. The common point set $ \mathcal{C} $ is defined as: 
\begin{equation}
\mathcal{C}(p_i^a, p_k^c) = \begin{cases} 
(p_i^a, p_j^b) \in \mathcal{M}_{i,j}, \\[1mm]
(p_j^{b'}, p_k^c) \in \mathcal{M}_{j,k}, \\[1mm]
|p_j^b - p_j^{b'}| < d_{NN},
\end{cases}
\label{common-point-eq}
\end{equation}
where $\mathcal{M}_{i,j}$ denotes matched point pairs between views $i$ and $j$, and $d_{NN}$ is a mutual nearest-neighbor distance threshold. Since identical physical points may not align exactly in pixel coordinates, a mutual nearest-neighbor search is performed in a KD-Tree: for each $p_j^b$, its nearest neighbor $p_j^{b'}$ is identified, and vice versa. If the distance between them is less than $d_{\text{NN}}$, they are considered the same point. The point propagation loss, similar to reprojection loss is formulated as:
\begin{equation}
L^{PPC}_{init} = \sum_{i,k} \| \pi(K, R_{ik}, Ray(p_i^a, z^a_i)) - p^c_k \|_2^2,
\end{equation}
where $R_{ik}$ denotes the relative pose from view $i$ to view $k$. Algorithm \ref{alg-2} describes the process of constructing the common point set $\mathcal{C}$.

\begin{algorithm}[!t]
\label{alg-111}
\caption{Common Point Set $ \mathcal{C} $}

\renewcommand{\algorithmicrequire}{\textbf{Input:}}
\renewcommand{\algorithmicensure}{\textbf{Output:}}

\begin{algorithmic}[1]
\small
\label{alg-2}

\REQUIRE Matching point pairs $\mathcal{M}_{i,j} = {(\mathbf{p}_i^a, \mathbf{p}_j^b)}$ between views $i$ and $j$, $\mathcal{M}_{j,k} = {(\mathbf{p}_j^{b'}, \mathbf{p}_k^c)}$ between views $j$ and $k$, pixel distance threshold $d_{NN}$;
\ENSURE Common point set $\mathcal{C} = {(\mathbf{p}_i^a, \mathbf{p}_k^c)}$ for indirect matching between views $i$ and $k$;

\STATE \textbf{Construct KD-Trees for View $j$}:
\STATE \hspace{0.5cm} Build KD-Tree $\mathcal{T}_j$ using all $\mathbf{p}_{j}^b$ from $\mathcal{M}_{i,j}$
\STATE \hspace{0.5cm} Build KD-Tree $\mathcal{T}'_j$ using all $\mathbf{p}_{j}^{b'}$ from $\mathcal{M}_{j,k}$

\STATE \textbf{Search for Mutual Nearest Neighbors}:
\STATE \hspace{0.5cm} Initialize empty set $\mathcal{C}$
\STATE \hspace{0.5cm} \textbf{for} each $\mathbf{p}_{j}^b \in \mathcal{M}_{i,j}$ \textbf{do}
\STATE \hspace{1.0cm} Find its nearest neighbor $\mathbf{p}_j^{b'}$ in $\mathcal{T}'_j$
\STATE \hspace{1.0cm} Find the nearest neighbor $\mathbf{p}_j^{b''}$ of that $\mathbf{p}_j^{b'}$ in $\mathcal{T}_j$
\STATE \hspace{1.0cm} \textbf{if} $\mathbf{p}_j^{b} = \mathbf{p}_j^{b''}$ \textbf{and} $|\mathbf{p}_j^b - \mathbf{p}_j^{b'}| < d_{NN}$ \textbf{then}
\STATE \hspace{1.5cm} Extract corresponding $\mathbf{p}_i^a$ from $\mathcal{M}_{i,j}$ and $\mathbf{p}_k^c$ from $\mathcal{M}_{j,k}$
\STATE \hspace{1.5cm} Add pair $(\mathbf{p}_i^a, \mathbf{p}_k^c)$ to $\mathcal{C}$
\STATE \hspace{1.0cm} \textbf{end if}
\STATE \hspace{0.5cm} \textbf{end for}

\STATE \textbf{return} $\mathcal{C}$
\label{alg-3}
\end{algorithmic}

\label{alg-4}

\vspace{-0.6mm}
\end{algorithm}

\textbf{Outlier Filtering Mechanism}: 
Mismatched pairs on object edges can have low reprojection error yet high depth differences, whereas those in untextured areas may have low depth differences but high reprojection error. Hence, the Outlier Filtering Mechanism jointly considers both metrics for the Reprojection Constraint:
\begin{equation}
\begin{cases}
\tau_{dy} = \tau_{base} + \alpha \cdot Sigmoid(2\hat{z} - 1), \\[2mm]
M_{i,j}^{depth} \leftarrow\mathcal{D}(z_{ij},z_{ji})= \frac{|z_{ij} - z_{ji}|}{\min(z_{ij},z_{ji})} \leq \tau_{dy}, \\[2mm]
 M_{ij}^{error} = E_{ij}^{RC} \leq \tau_{RC} ,\\[2mm]
M_{i,j}^{RC\text{-}dep} = M_{ij}^{error} \odot M_{ji}^{error} \odot M_{i,j}^{depth},
\end{cases}
\label{rc-mask-eq}
\end{equation}
where the dynamic threshold $\tau_{dy}$ is implemented using a Sigmoid function to enforce stricter depth consistency constraints for nearby points and more relaxed tolerances for distant points; $\tau_{base}$ denotes the base threshold, $\alpha$ is a scaling factor, and $\hat{z}$ represents the normalized depth value; $z_{ij}$ denotes the ray depth of view $i$ in the view pair $i$-$j$; $M_{i,j}^{depth}$ is the depth consistency mask, obtained by comparing the relative depth difference with $\tau_{dy}$; $E_{ij}^{RC}$ denotes the reprojection error between views $i$ and $j$; $\tau_{RC}$ is a fixed reprojection error threshold; $M_{ij}^{error}$ is the reprojection error mask; $M_{i,j}^{RC\text{-}dep}$ is the final valid matching mask that combines bidirectional reprojection errors and depth consistency.
Based on this mechanism, matched point pairs are filtered if either the reprojection error exceeds the predefined threshold or the relative depth difference is too large.

The same filtering principle is extended to multi-view chains for the Point Propagation Constraint. The co-visible point set through the bridge view $j$ is filtered as:
\begin{equation}
\begin{cases}
M^{PPC}_{jik}\leftarrow  \mathcal{C}(p_i^a, p_k^c),M_{i,k}^{depth}\leftarrow\mathcal{D}(z_{ij},z_{kj}), \\[2mm]
M_{jik}^{PPC\text{-}dep} =M^{PPC}_{jik} \odot M_{i,k}^{depth},
\end{cases}
\label{pcc-mask-q}
\end{equation}
where $M_{jik}^{PPC}$ denotes the initial matching mask for the propagation path $i\rightarrow j_{bridge}\rightarrow k$,  obtained from Eq.~\ref{common-point-eq}, and $M_{i,k}^{depth}$ is obtained from Eq.~\ref{rc-mask-eq}. $M_{jik}^{PPC\text{-}dep}$ represents the point propagation mask after applying depth-based filtering.

Additionally, two enhancement operations are applied: point propagation relations with a common point count below a predefined threshold are filtered out. In 360° unbounded scenes, point propagation relations are constructed based on the nearest-neighbor distances computed between different views. These two operations reduce unnecessary computations and significantly accelerate processing.

\begin{figure}[t]
\centering

\includegraphics[width=3.4in]{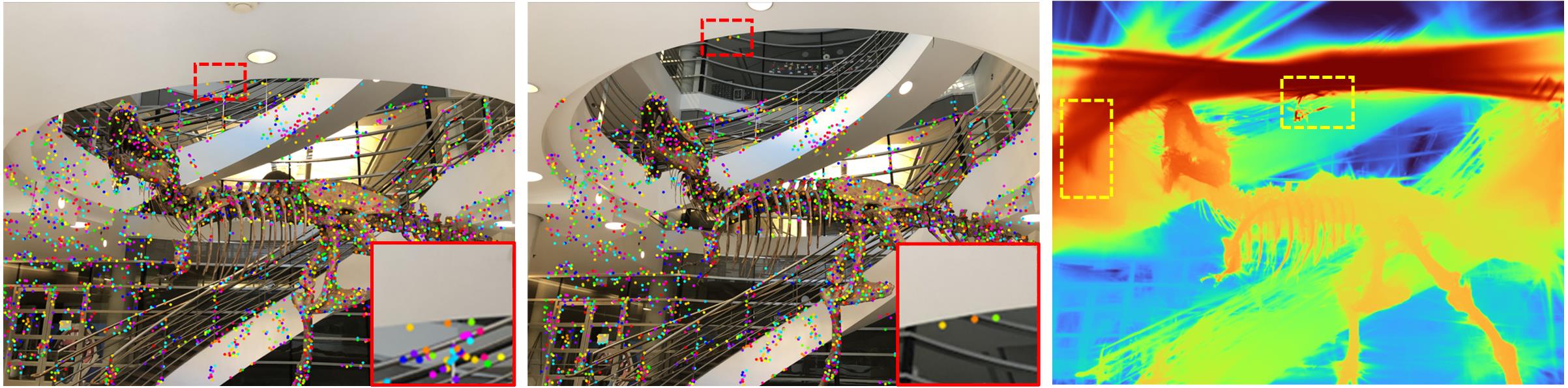}

\begin{center}
\vspace{-1mm}
\footnotesize  
 (a) 
Matching pair images (left, middle) and depth map(right) of SCGaussian.
\end{center}

\vspace{1mm}
\includegraphics[width=3.4in]{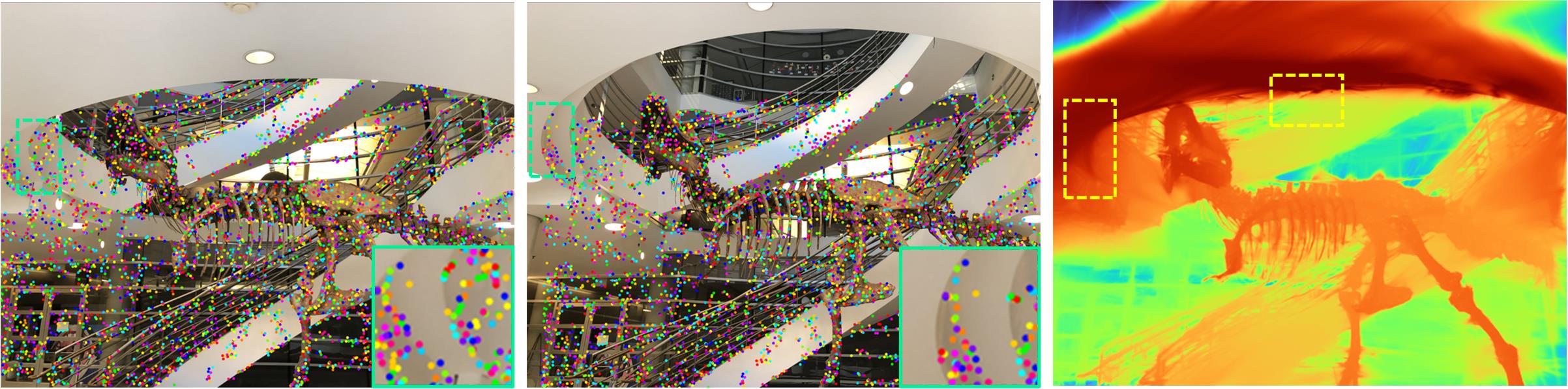}
\begin{center}
\footnotesize
(b) Matching pair images (left, middle) and depth map (right) of HBSplat.
\end{center}

\vspace{-2mm}
\caption{Visual comparison of matching pair images and depth maps. (a) and (b) respectively show the matching pair images (left, middle) and corresponding depth maps (right) for SCGaussian and HBSplat. SCGaussian relies on the RANSAC algorithm for outlier filtering, whereas HBSplat employs Outlier Filtering Mechanism. The comparison demonstrates that HBSplat more effectively removes outliers while preserving a greater number of valid points. }
\vspace{-1.2mm}
\label{matching-f}
\end{figure}

\subsubsection{Scene Training Stage}
In the scene training stage, we employ refined depth constraints, converting filtered matching points into spatial points and then into Gaussian primitives. To ensure robustness, the training stage includes three loss functions to constrain the rendered depth as follows: 

\textbf{Reprojection \& Point Propagation Constraint}: Inspired by SCGaussian\cite{ref33}, we apply Bilinear Sampling to the rendered depth map along rays, computing reprojection and point propagation errors from training views to other views to enforce spatial structure consistency. 
\begin{equation}\label{filter-eq}
\dot{p} = M \odot p,
\end{equation}
\begin{equation}\label{rc-train}
L^{RC}_{train} = \sum_{i,j} \| \pi(K, R_{ij}, Ray(\dot{p}^a_i, Bil(D_i))) - \dot{p}^b_j \|_2^2,
\end{equation}
\begin{equation}\label{ppc-train}
L^{PPC}_{train} = \sum_{i,k} \| \pi(K, R_{ik}, Ray(\dot{p}^a_i, Bil(D_i)) - \dot{p}^c_k \|_2^2,
\end{equation}
where $\dot{p}$ denotes the pixel point $p$ after filtering, which is performed using either the mask $M_{i,j}^{RC\text{-}dep}$ or $M_{jik}^{PPC\text{-}dep}$. $ D_i $ is the rendered depth map of view $i$, $Ray(\cdot)$ denotes the ray function that generates spatial points, $\pi(\cdot)$ denotes the projection function, $Bil(\cdot)$ denotes bilinear sampling. 

\textbf{Smoothness Constraint}: The Total Variation (TV) loss is applied to promote spatial smoothness in the rendered depth map:
\begin{equation}
L_{TV} = \sum_{u,v} \left( | \nabla_x D_i(u,v) | + | \nabla_y D_i(u,v) | \right),
\end{equation}
where $u,v$ are pixel coordinates, $D_i$ is the rendered depth map of view $i$, and $\nabla_x$ and $\nabla_y$ denote the horizontal and vertical gradient operators, respectively.

The final depth loss functions for the Initial Depth Estimation Stage and the Scene Training Stage are defined as follows:
\begin{equation}
L_{init} = L^{RC}_{init} + L^{PPC}_{init},
\end{equation}
\begin{equation}
\label{train_d_loss}
L_{train} = L^{RC}_{train} + L^{PPC}_{train}+L_{TV}.
\end{equation}

Figure \ref{matching-f} presents a comparative visualization between the results without and with the Outlier Filtering Mechanism, along with a comparison of the depth maps generated after applying the Hybrid-Loss.

\begin{figure*}[!t]
\centering
\subfloat{\includegraphics[width=6.7in]{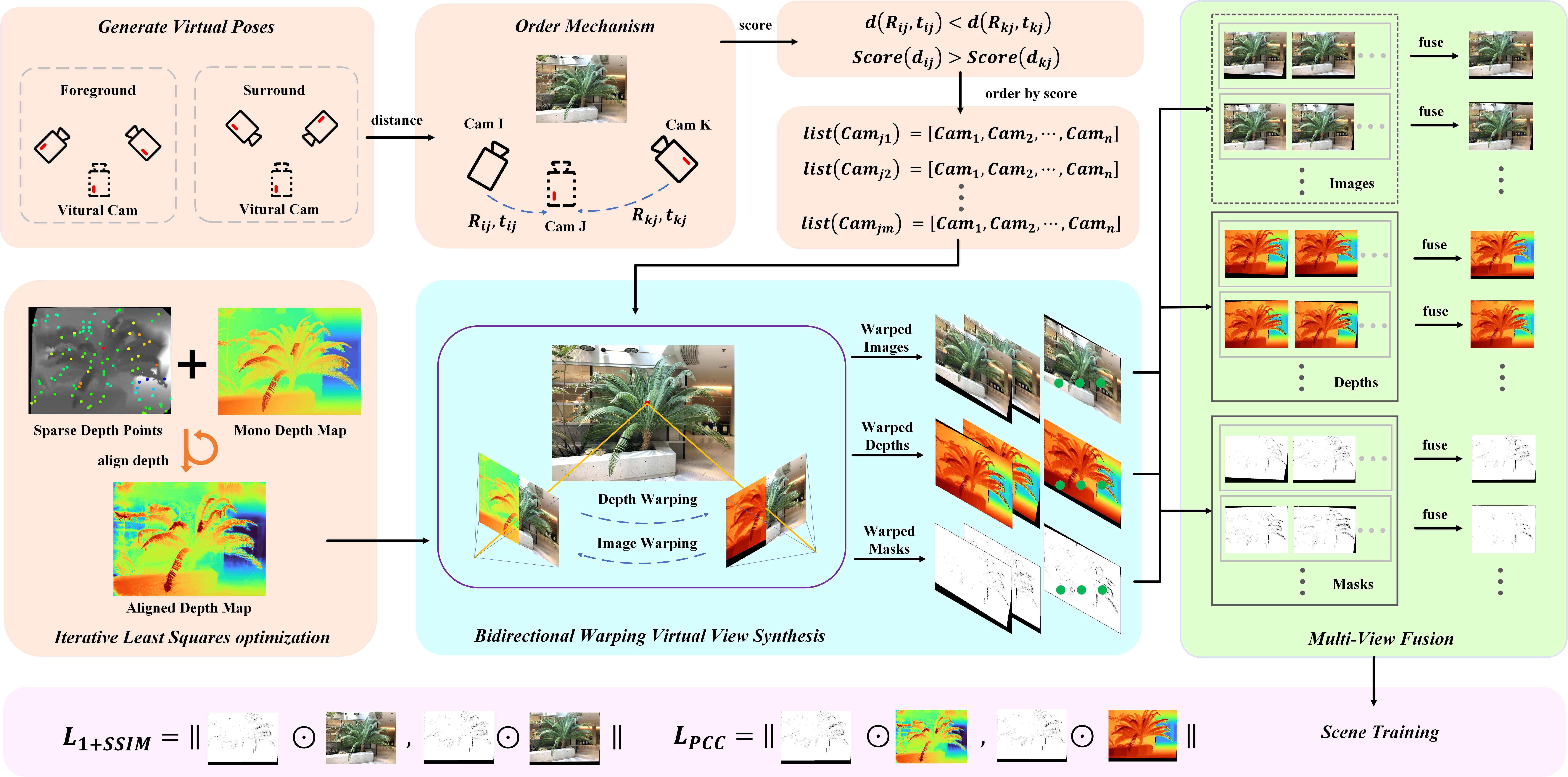}%
\label{fig_first_case}}
\hfil
\caption{Bidirectional Warping Virtual View Synthesis Pipeline. The monocular depth maps from the real views are aligned with sparse depth points via least-squares optimization. Depth Warping generates virtual depth maps, filling holes. Image Warping samples real views to create virtual views. Distance scores between real and virtual views are computed, warping multiple optimal real views to a single virtual view. The nearest-neighbor virtual view serves as the base for Multi-View Fusion. The 3D Gaussian scene is reconstructed using gradient-domain loss and Pearson Correlation Coefficient (PCC) loss constraints.}
\label{bidirectional-f}
\end{figure*}

\subsection{Bidirectional Warping Virtual View Synthesis}
\subsubsection{Virtual View Generation Strategy}
In sparse-view scenarios, particularly with as few as 3 images, 3DGS rendering often suffers from overfitting and severe geometric distortions due to insufficient multi-view constraints, especially when input views are spatially clustered. To mitigate this, we propose generating multiple virtual views via Depth-Image-Based Rendering (DIBR) to impose stronger photometric and geometric constraints during 3DGS optimization. DIBR, also known as 3D Image Warping, is a core computer graphics technique that projects reference images into 3D space using depth maps and reprojects 3D points onto a virtual camera’s image plane. Two primary warping strategies exist: (1) Forward Warping, which is computationally simple but often produces holes and disocclusions, leading to less effective synthesis; and (2) Backward Warping, which provides higher-quality results but requires the depth map of the target view. Our approach integrates both strategies through Bidirectional Warping, effectively combining their advantages. Quantitative comparison between Forward and Bidirectional Warping is provided in Table \ref{tab:warping_comparison}.

The Bidirectional Warping process operates in two sequential stages—Depth Warping followed by Image Warping—to effectively overcome the limitations of pure Forward Warping. Specifically, the method first warps and interpolates depth values to generate a complete virtual depth map, then performs reverse warping for photometrically consistent color interpolation. This approach yields virtual views with significantly reduced artifacts such as graininess and blurring while maintaining computational efficiency. The overall pipeline is illustrated in Figure \ref{bidirectional-f}, and the warping process is formulated as follows:
\begin{equation}
\begin{cases}
p_{vir} = K_{vir} \cdot R_{vir} \cdot R_{src}^{-1} \cdot (D_{src} \cdot K_{src}^{-1} \cdot p_{src}), \\
D_{vir}(p_{vir}) = Fill(Bil(D_{src}(p_{src}))),
\end{cases}
\end{equation}
\begin{equation}
\begin{cases}
p_{src} = K_{src} \cdot R_{src} \cdot R_{vir}^{-1} \cdot (D_{vir} \cdot K_{vir}^{-1} \cdot p_{vir}), \\

I_{vir}(p_{vir}) = Bil(I_{src}(p_{src})),
\end{cases} 
\end{equation}
where $p_{src}$ and $p_{vir}$ denote pixels in the real view and the virtual view, respectively, $D_{vir}$ is the virtual depth map generated through Depth Warping, and $I_{vir}$ is the synthesized virtual image obtained via Image Warping. The source depth map $D_{src}$ is estimated using monocular depth estimator, with scale consistency enforced via least-squares optimization. The $Fill(\cdot)$ operator handles hole completion in the warped depth map, $Bil(\cdot)$ denotes bilinear sampling.

\begin{table}[!t]
    \centering
    \renewcommand\arraystretch{1.3}
    \caption{Comparison between Forward Warping and Bidirectional Warping metrics.}
    
    \begin{tabular}{c|c|c|c|c|c|c}
    \hline
    \multirow{2}{*}{Scene} & \multicolumn{3}{c|}{Forward Warping} & \multicolumn{3}{c}{Bidirectional Warping} \\
    \cline{2-7}
    & PSNR$\uparrow$ & SSIM$\uparrow$ & LPIPS$\downarrow$ & PSNR$\uparrow$ & SSIM$\uparrow$ & LPIPS$\downarrow$ \\
    \hline
    Fern & 22.58 & 0.751 & 0.175 & 22.67 & 0.752 & 0.175 \\
    Leaves & 18.07 & 0.658 &  0.238 & 18.14 & 0.659 & 0.237 \\
    Room & 22.43 & 0.868 & 0.142 & 22.53 & 0.868 & 0.141\\
    \hline
    \end{tabular}
    \label{tab:warping_comparison}

\end{table}

\subsubsection{Scale-Agnostic Monocular Depth Recovery} 
Virtual view synthesis relies on accurate depth maps from the real views. The scale and offset of scale-agnostic monocular depth maps are recovered via an iterative least-squares optimization:
\begin{equation}\label{equ}
s', b' = \arg\min_{s,b} \sum_{{z} \in D_{SP}} \| M_{RC} \odot D_{SP}(z) - D'(s,b) \|_2^2,
\end{equation}
where $D'$ is obtained through the linear transformation $D' = s \cdot D_{mono} + b$, $D_{SP}$ is sparse depth points generated from robust camera-space depths $z$ obtained through Hybrid-Loss Depth Estimation, and $M_{RC}$ is the reprojection error mask.
The training view's depth map with the recovered scale $s'$ and offset $b'$ is given by:
\begin{equation}\label{equ}
D_{src} = s' \cdot D_{mono} + b'.
\end{equation}

\subsubsection{Virtual View Selection \& Fusion}
\textbf{Virtual Pose Selection via Pose Scoring:}
To generate high-quality virtual views, a scoring function is proposed to prioritize source real views that are geometrically close to the target virtual pose. The Nearest Neighbor Score for each candidate pose is computed based on both translational distance and rotational difference, with higher scores assigned to closer poses with more similar orientations.

\begin{figure}
\centering
\includegraphics[width=3.4in]{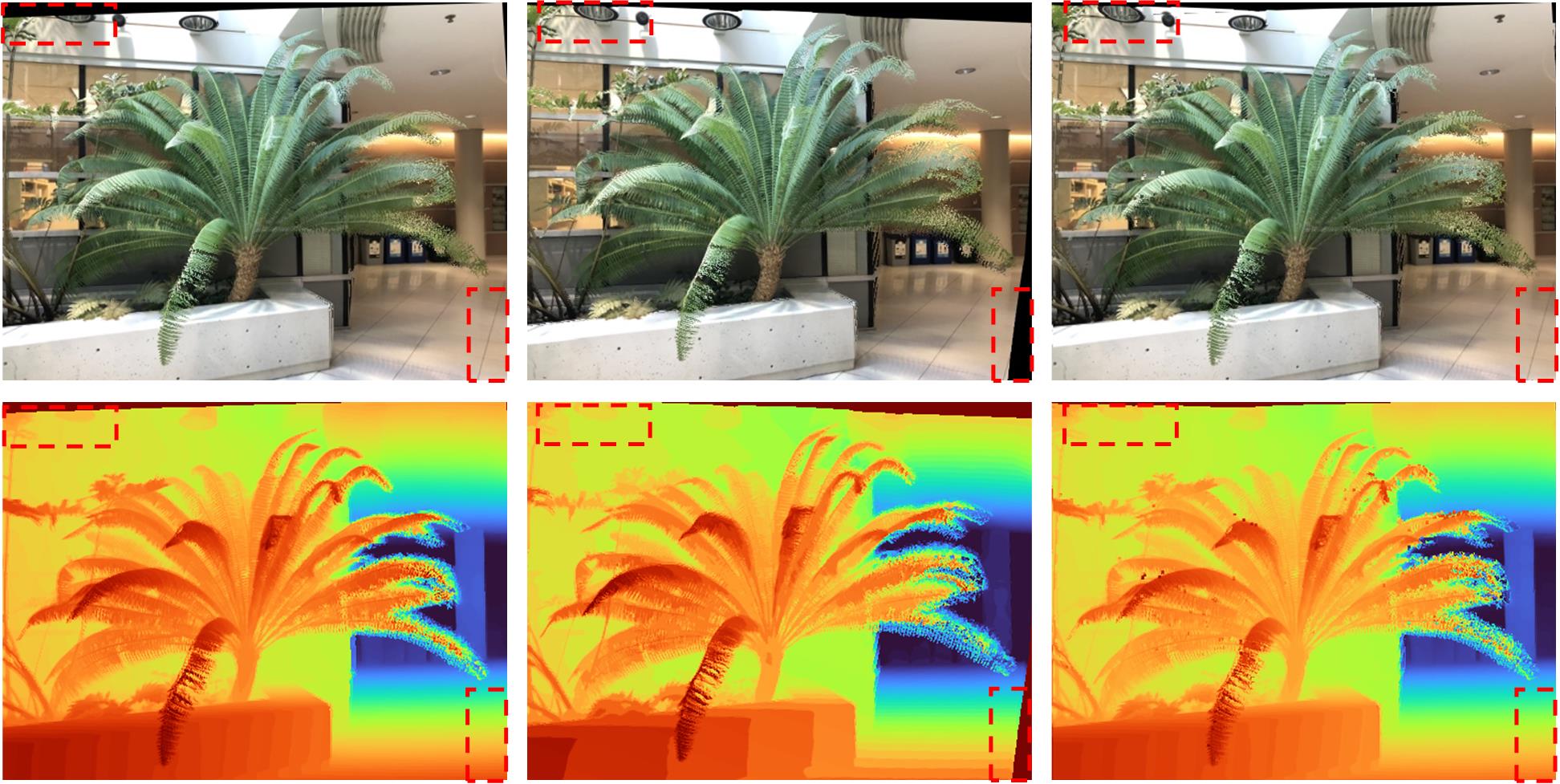}
\caption{Multi-View Fusion. Fusing the left, middle virtual views produces the result shown on the right.}
\label{fusion-f}
\end{figure}

\textbf{Multi-View Fusion:} Since a virtual view generated from a single source view may contain occlusions or missing regions, a fusion strategy aggregates information from multiple sources. Specifically, the top-$k$ virtual views rendered at the same target pose are fused; these views are generated from source views selected based on the highest pose similarity scores. The view from the closest source serves as a base, while others are blended to expand visible regions. The complete fusion process is formulated as follows:
\begin{equation}
I'_{vir} = Fuse_k(I_{vir} | Order(Score(R_{src}, R_{vir}))),
\end{equation}
where $Score(\cdot)$ denotes the Nearest Neighbor Score quantifying pose similarity, $Order(\cdot)$ sorts source poses in descending order based on their $Score$ values, indicating their proximity to the target virtual pose, $Fuse_k(\cdot)$ represents the operation of fusing the top-$k$ virtual views, and $I'_{vir}$ is the final fused virtual view. Figure \ref{fusion-f}
illustrates the fusion result.

The virtual depth map and occlusion mask corresponding to the virtual view are also generated through Bidirectional Warping and the Selection \& Fusion operations, as illustrated in Figure
\ref{bidirectional-f}.

\subsubsection{Loss function} 
The loss function for virtual view supervision is defined as follows:
\begin{equation} 
\label{vir_color}
\hspace{-0.5mm}
\begin{cases}
L_{grad}=L_1(G(M) \odot G(I_{ren}), G(M) \odot G(I'_{vir})), \\[1mm]
L^{vir}_{color}=\alpha  L_{grad} + \beta L_{ssim}(M \odot I_{ren}, M \odot I'_{vir}) ,
\end{cases}
\end{equation}
\begin{equation} 
\label{vir_depth}
L^{vir}_{depth}=PCC(M \odot D_{ren}, M\odot D_{vir}),
\end{equation}
where $M$ is the virtual view mask, $G(\cdot)$ computes the image gradient domain, $\alpha$ and $\beta$ are weighting coefficients satisfying $\alpha + \beta = 1$, and $\text{PCC}(\cdot)$ denotes the Pearson Correlation Coefficient operation.
The gradient-domain loss $L_{\text{grad}}$ enhances sensitivity to edges and structural details by emphasizing high-frequency information, while maintaining robustness to absolute intensity variations. The Pearson Correlation Coefficient for depth loss measures linear dependence between rendered and virtual depths, effectively capturing structural consistency while being invariant to global scale shifts.

\subsection{Occlusion-Aware Reconstruction}
In NVS, models are typically optimized to reconstruct only the visible regions present in the training views. However, synthesizing views beyond the original training set often reveals occluded areas. This issue is exacerbated under sparse-view conditions, where limited coverage increases the probability of exposing unseen regions in novel views.

\begin{figure}[t!]
\centering
\includegraphics[width=3.4in]{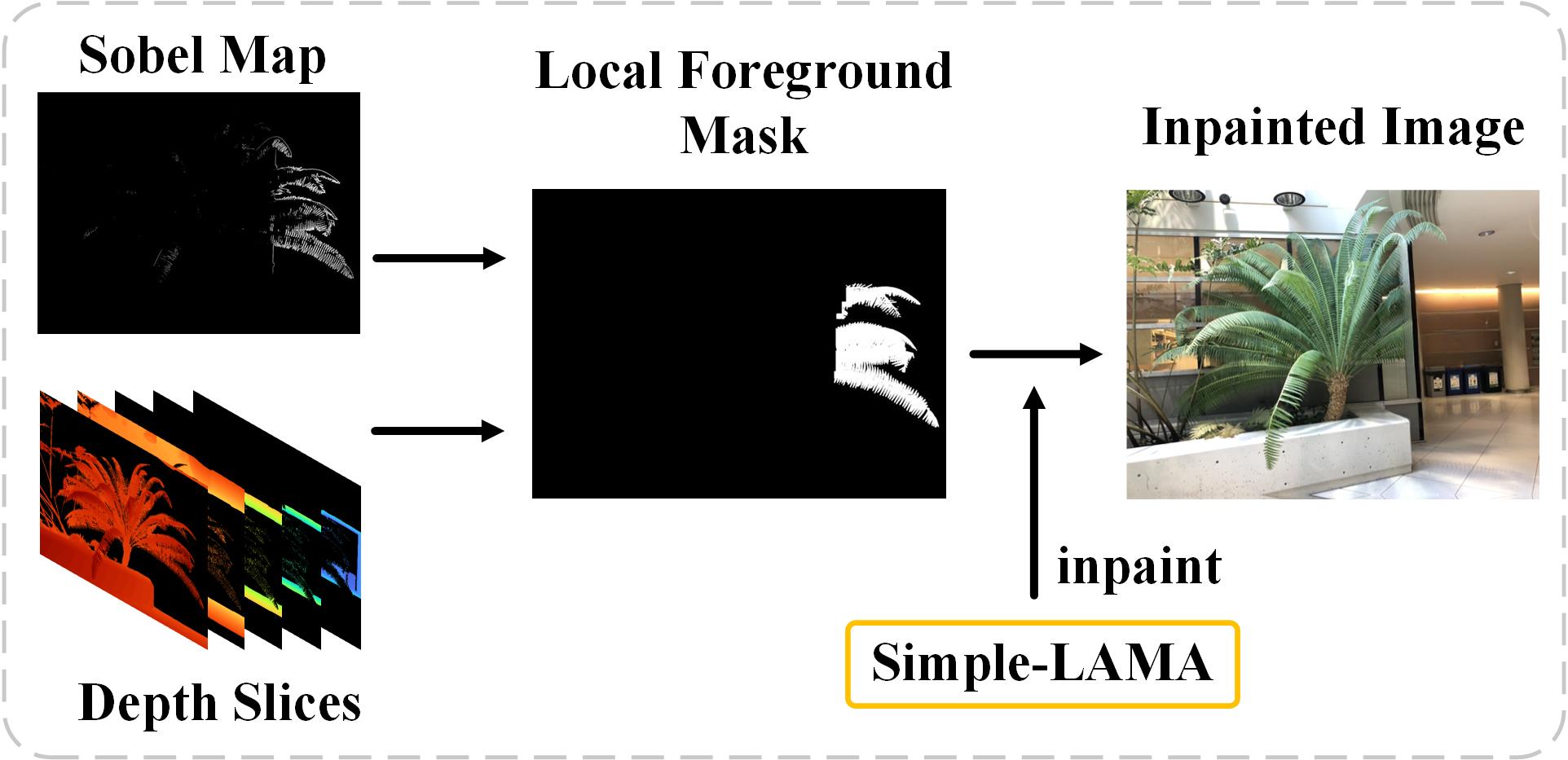}
\caption{Occlusion-Aware Reconstruction Pipeline. Local foreground mask, generated by utilizing the sobel map and depth slices, identifies the inpainting regions and guides the Simple-LAMA model to perform the inpainting.}
\label{pipeline3-f}
\end{figure}

Regions exhibiting substantial depth differences between foreground and background are particularly prone to occlusions. Based on this, an Occlusion-Aware Reconstruction method grounded in depth differences is proposed, as illustrated in Figure \ref{pipeline3-f}. This component is suited for reconstructing smaller occluded regions. First, edge gradients of the depth map are computed using the Sobel operator to extract candidate regions $a$ with large foreground-background depth disparities. Since such edge regions often fail to form continuous occlusion boundaries, the monocular depth map $D_{\text{mono}}$ is partitioned into $n$ discrete depth layers. The depth slice $b$ where pixels from region $a$ are predominantly distributed is identified to generate an initial foreground mask. To refine the mask and exclude foreground regions with insignificant depth variations, the minimum bounding rectangle $c$ of region $a$ is extracted. This rectangle is used to crop content from depth slice $b$, producing a local foreground mask emphasizing regions with pronounced depth differences:
\begin{equation}
M_{local\_FG} = \mathcal{F}(Soble(D_{mono}), Slice(D_{mono})),
\end{equation}
where $D_{mono}$ denotes the monocular depth map, $Sobel(\cdot)$ represents the Sobel edge detection operator, $Slice(\cdot)$ refers to the depth slicing operation, and $\mathcal{F}$ denotes the function that extracts the local foreground mask $M_{local\_FG}$.

The Simple-LAMA~\cite{ref40} inpainting model is subsequently employed to generate plausible completion results. This model uses the $M_{local\_FG}$ along with the training views as input, where the mask specifically identifies regions that require completion:
\begin{equation}
I_{inp} = Lama(M_{local\_FG}, I),
\end{equation}
where $I_{inp}$ denotes the image after inpainting. To enforce geometric consistency and refine the reconstruction, the inpainting image $I_{inp}$ is used to supervise the background Gaussian splatting process. The background point set is first isolated by filtering out foreground regions using the occlusion-aware mask:
\begin{equation}
P_{BG} = P \odot  \lnot(M_{local\_FG} \odot  M_{D\_slice}),
\end{equation}
\begin{equation}
\mathcal{S}(t) = \begin{cases} 
Train(I_{inp}, P_{BG}),& if \  t \leq t_{BG}, \\
Train(I,\ P), & if \ t > t_{BG},
\end{cases}
\end{equation}
where $M_{D\_slice}$ is a depth slice mask that identifies regions with depth values less than the current slice threshold, $P$ denotes the original 3D point cloud, and $P_{BG}$ represents the point cloud after excluding points in the occluded foreground regions. The background Gaussians are optimized exclusively using $I_{inp}$ as supervision for the first $t_{BG}$ iterations, after which the entire set of Gaussians is trained to reconstruct the complete scene. Ablation will demonstrate the effectiveness of the proposed Occlusion-Aware Reconstruction.

\subsection{Total Loss Function}
The overall optimization objective combines photometric, geometric, and virtual view consistency terms, defined as:
\begin{equation}
L_{col}(I) = \lambda L_1(I_{ren}, I) + (1-\lambda )L_{ssim}(I_{ren}, I),
\end{equation}
\begin{equation}
L_{total} = L_{col}(I) + L_{dep}(D) + L_{col}(I_{vir}) + L_{dep}(D_{vir}),
\end{equation}
where $L_{dep}(D)$ denotes the depth loss as defined in Eq.~\ref{train_d_loss}, $L_{col}(I_{vir})$ denotes the virtual image color loss as defined in Eq.~\ref{vir_color}, and $L_{dep}(D_{vir})$ denotes the virtual depth loss as defined in Eq.~\ref{vir_depth}). The weight $\lambda$ is set to 0.8 in all experiments.

\section{Experiment}
\subsection{Setup}

\subsubsection{Datasets} 
HBSplat is evaluated on five widely adopted public datasets: LLFF, IBRNet, Blender, DTU, and Tanks\&Temples. These datasets cover forward-facing, synthetic, object-centric, and 360° unbounded scenes, ensuring a comprehensive assessment of robustness across diverse settings.
For LLFF and IBRNet (forward-facing scenes), we follow the protocol of DNGaussian~\cite{ref22}: one every 8 images is held out for testing, while 3 training views are uniformly sampled from the remainder. Evaluation is performed over LLFF scenes with 8×, 4× downsampling rate and IBRNet scenes with 2× downsampling rate.
For Blender (synthetic objects), models are trained on 8 views and evaluated on 25 uniformly sampled test images with 2× downsampling rate.
For DTU (object-centric scenes), we adopt the setup from SCGaussian~\cite{ref36}, using provided object masks during evaluation to exclude background regions and focus on the target object. 4× downsampling rate is applied.
For Tanks\&Temples (360° scenes), 6 scenes are selected. Training uses 24 views, training and testing split strategy follows the same protocol as used in LLFF (one every 8 images) with 4× downsampling rate.

\begin{figure*}[!t]
\centering
\subfloat{\includegraphics[width=7in]{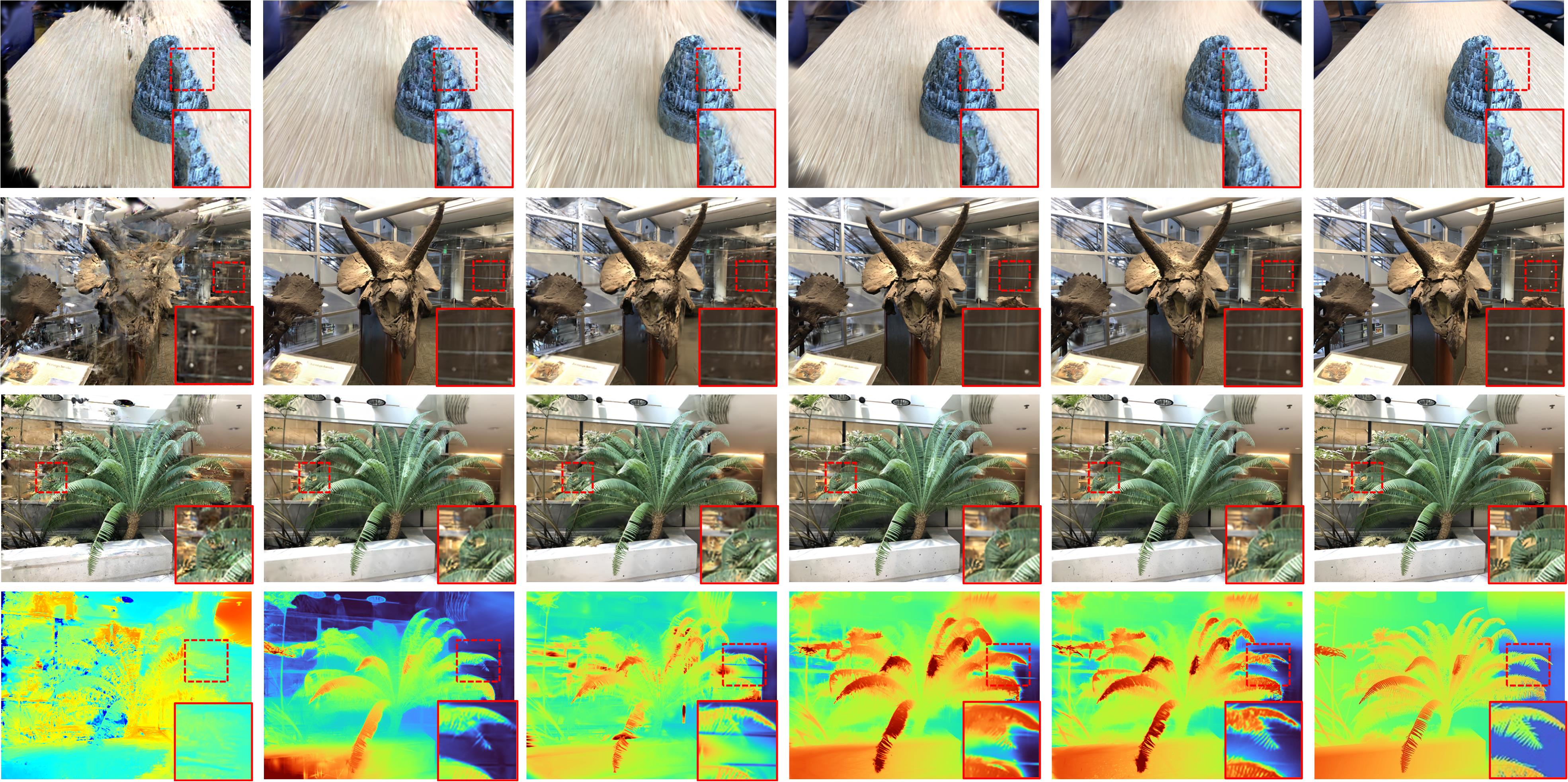} 
\label{fig_first_case}}
\hfil

\begin{center}
\vspace{-1mm}
\footnotesize  
 3DGS\hspace{23mm}
 SID\hspace{22mm}
 MCGS\hspace{18mm}
 SCGaussian\hspace{14mm}
 HBSplat(Ours)\hspace{19mm}
 GT
\end{center}
\vspace{-2mm}
\caption{Qualitative comparison on LLFF dataset under 3 training views. The rendering results of our HBSplat are more accurate and display finer details.}
\label{ll1}
\end{figure*}

\begin{figure*}
\centering
\subfloat{\includegraphics[width=7in]{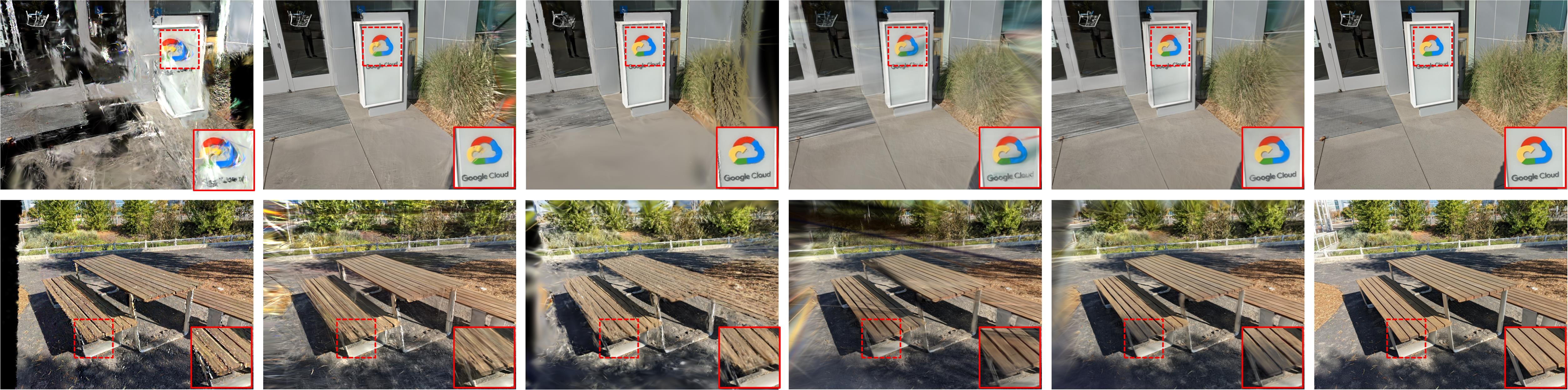} 
\label{fig_first_case}}
\hfil

\begin{center}
\vspace{-1mm}
\footnotesize  
 3DGS\hspace{21mm}
 FSGS\hspace{18mm}
 DNGaussian\hspace{15mm}
 SCGaussian\hspace{15mm}
 HBSplat(Ours)\hspace{19mm}
 GT
\end{center}
\vspace{-2mm}
\caption{Qualitative comparison on IBRNet dataset with 3 training views. The rendering results of HBSplat method are more accurate and display finer details.}

\label{ib-p}
\end{figure*}

\subsubsection{Evaluation Metrics} 
Rendering quality is evaluated using PSNR, SSIM, and LPIPS. In addition, following DNGaussian~\cite{ref22}, the Average (AVG) is computed as a composite metric, defined as the geometric mean of  MSE ($10^{-\text{PSNR}/10}$), $\sqrt{1-\text{SSIM}}$, and LPIPS. Efficiency is assessed via frames per second (FPS) and average training time. Final dataset-level metrics are reported as the average across all scenes.

\subsubsection{Baselines} 
HBSplat is compared against several excellent sparse-view NVS jobs, including NeRF-based and 3DGS-based approaches. NeRF-based baselines include FreeNeRF\cite{ref11} and SparseNeRF\cite{ref9}. 3DGS-based baselines include vanilla 3DGS\cite{ref2}, FSGS\cite{ref20}, SID\cite{ref39}, DNGaussian \cite{ref22}, MCGS\cite{ref32}, and SCGaussian \cite{ref36}. For NeRF-based methods, we report their best quantitative results from the respective papers. To ensure a fair comparison, we use public code with consistent training-testing splits and a unified base environment with core dependencies fixed, including PyTorch and CUDA.

\subsubsection{Implementation Details} 
HBSplat is built upon the 3DGS framework\cite{ref2}. Camera poses are estimated using COLMAP across all views for robustness. To ensure consistent comparison, similar to SCGaussian, we employed the GIM-DKM\cite{ref37} model for dense matching to extract matching points per training view. Besides, we use ML-depth-pro~\cite{ref38} for monocular depth estimation, which excels in capturing fine details. Both depth estimation and scene training are iterated 2,000 times. Initialize and generate Gaussian primitives after filtering outliers. Bidirectional Warping generates 100 virtual views; one virtual view iteration is performed for every 5 real view iterations. The Occlusion-Aware Reconstruction applies a depth difference mask to prioritize rendering of occluded region Gaussians in the first 1,000 iterations, followed by training all Gaussians. Gaussian cloning and pruning follow the vanilla 3DGS settings. All experiments are conducted on a single NVIDIA RTX 3090 GPU. For detailed configuration, please refer to the publicly available code.

\begin{table*}[t]
    \centering
    \caption{Quantitative comparisons on the LLFF (1/8, 1/4 resolution), IBRNet (1/2 resolution) datasets with 3 training views. Top-3 entries are highlighted: red (1st), orange (2nd), yellow (3rd).}
    
    \renewcommand\arraystretch{1.3}
    \begin{tabular}{ 
        l | 
        >{\centering\arraybackslash}p{0.75cm} |>{\centering\arraybackslash}p{0.75cm} |>{\centering\arraybackslash}p{0.75cm} |>{\centering\arraybackslash}p{0.75cm} |
        >{\centering\arraybackslash}p{0.75cm} |>{\centering\arraybackslash}p{0.75cm} |>{\centering\arraybackslash}p{0.75cm} |>{\centering\arraybackslash}p{0.75cm} |
        >{\centering\arraybackslash}p{0.75cm} |>{\centering\arraybackslash}p{0.75cm} |>{\centering\arraybackslash}p{0.75cm} |>{\centering\arraybackslash}p{0.75cm} 
    }
    
    \hline
    \multicolumn{1}{c|}{\multirow{2}{*}{Method}} & \multicolumn{4}{c|}{1/8 LLFF} & \multicolumn{4}{c|}{1/4 LLFF} & \multicolumn{4}{c}{1/2 IBRNet}\\
    \cline{2-13}
    & PSNR$\uparrow$ & SSIM$\uparrow$ & LPIPS$\downarrow$ & AVG$\downarrow$ & PSNR$\uparrow$ & SSIM$\uparrow$ & LPIPS$\downarrow$ & AVG$\downarrow$ & PSNR$\uparrow$ & SSIM$\uparrow$ & LPIPS$\downarrow$ & AVG$\downarrow$\\ 

    \hline    FreeNeRF\pub{CVPR23}\cite{ref11}    & 19.63 & 0.612 & 0.308 & 0.134     & 18.73 & 0.562 & 0.384 & 0.169     & 19.76 & 0.588 & 0.333 & 0.135 \\
    SparseNeRF\pub{ICCV23}\cite{ref9}   & 19.86 & 0.624 & 0.328 & 0.127     & 19.07 & 0.564 & 0.401 & 0.168     & 19.90 & 0.593 & 0.364 & 0.137 \\

    \hline
    3DGS\pub{SIGGRAPH23}\cite{ref2}         & 15.42 & 0.383 & 0.463 & 0.218     & 13.28 & 0.350 & 0.486 & 0.264     & 17.20 & 0.556 & 0.355 & 0.165 \\
    FSGS\pub{ECCV24}\cite{ref20}        & 20.27 & 0.697 & 0.206 & 0.102     & \cellcolor{myyellow}19.70 & \cellcolor{myyellow}0.667 & \cellcolor{myyellow}0.265 & \cellcolor{myyellow}0.117     & 19.67 & 0.605 & 0.306 & 0.127 \\
    SID\pub{ICASSP25}\cite{ref39}         & \cellcolor{myyellow}20.40 & \cellcolor{myyellow}0.701 & \cellcolor{myyellow}0.215 & \cellcolor{myyellow}0.102     & 19.08 & 0.635 & 0.336 & 0.135     & 19.44 & 0.599 & 0.360 & 0.137 \\
    DNGaussian\pub{CVPR24}\cite{ref22}  & 19.12 & 0.591 & 0.294 & 0.132     & 18.23 & 0.575 & 0.386 & 0.155     & 18.14 & 0.554 & 0.415 & 0.161 \\
    MCGS\pub{TPAMI25}\cite{ref32}        & 20.32 & 0.700 & 0.219 & 0.103     & 19.63 & 0.663 & 0.292 & 0.122     & \cellcolor{myyellow}21.02 & \cellcolor{myyellow}0.674 & \cellcolor{myyellow}0.282 & \cellcolor{myyellow}0.109 \\
    SCGaussian\pub{NeurIPS24}\cite{ref36}  & \cellcolor{myorange}20.73 & \cellcolor{myorange}0.725 & \cellcolor{myorange}0.196 & \cellcolor{myorange}0.095     & \cellcolor{myorange}20.03 & \cellcolor{myorange}0.683 & \cellcolor{myorange}0.266 & \cellcolor{myorange}0.114     & \cellcolor{myorange}21.47 & \cellcolor{myorange}0.689 & \cellcolor{myorange}0.275 & \cellcolor{myorange}0.103 \\ 

    \hline
    HBSplat(Ours)              & \cellcolor{myred}21.13 & \cellcolor{myred}0.735 & \cellcolor{myred}0.189 & \cellcolor{myred}0.090     & \cellcolor{myred}20.30 & \cellcolor{myred}0.693 & \cellcolor{myred}0.256 & \cellcolor{myred}0.109     & \cellcolor{myred}22.19 & \cellcolor{myred}0.708 & \cellcolor{myred}0.249 & \cellcolor{myred}0.093  \\ 
    \hline

    \end{tabular}
    \label{tab:ll}
    \vspace{-3mm}
\end{table*}

\begin{table*}
    \centering
    \caption{Quantitative comparisons on the NeRF-LLFF dataset with 2,3,5 training views. The best and second-best entries are marked in red and orange, respectively.}
    \vspace{-1mm}\renewcommand\arraystretch{1.1}
    \setlength{\tabcolsep}{3.7pt}
    \begin{tabular}{c|c|cccc|cccc|cccc}

    \hline
    \multirow{2}{*}{Scene}
    & \multirow{2}{*}{Method}
    & \multicolumn{4}{c|}{2-view}
    & \multicolumn{4}{c|}{3-view}
    & \multicolumn{4}{c}{5-view} \\
    \cline{3-14}
    & & PSNR $\uparrow$ & SSIM $\uparrow$ & LPIPS $\downarrow$ & AVG $\downarrow$
      & PSNR $\uparrow$ & SSIM $\uparrow$ & LPIPS $\downarrow$ & AVG $\downarrow$
      & PSNR $\uparrow$ & SSIM $\uparrow$ & LPIPS $\downarrow$ & AVG $\downarrow$ \\
    \hline
    \multirow{3}{*}{Fern}
    & 3DGS\pub{SIGGRAPH23} & 14.07 & 0.306 & 0.509 & 0.255
          & 15.80 & 0.348  & 0.502 & 0.220
          & 25.02 & 0.647 & 0.255 & 0.078 \\
    & SCG\pub{NeurIPS24} & 19.38 & 0.627 & 0.317 & 0.133
          & 22.15 & 0.741 & 0.175 & 0.084
          & 24.20 & 0.824 & 0.120 & 0.057 \\
    & HBSplat(Ours) & \textbf{20.09} & \textbf{0.658} & \textbf{0.258} & \textbf{0.115}
          & \textbf{22.73} & \textbf{0.753} & \textbf{0.173} & \textbf{0.078}
          & \textbf{24.52} & \textbf{0.832} & \textbf{0.110} & \textbf{0.054} \\
    \hline
    \multirow{3}{*}{Flower}
    & 3DGS\pub{SIGGRAPH23} & 16.15 & 0.370 & 0.437 & 0.203
          & 16.37 & 0.445  & 0.437 & 0.195
          & 26.10 & 0.782 & 0.146 & 0.055 \\
    & SCG\pub{NeurIPS24} & 19.84 & 0.608 & 0.307 & 0.134
          & 21.66 & 0.690 & \textbf{0.217} & 0.103
          & 25.86 & 0.858 & 0.097 & 0.045 \\
    & HBSplat(Ours) & \textbf{20.10} & \textbf{0.630} & \textbf{0.274} & \textbf{0.126}
          & \textbf{21.70} & \textbf{0.693} & 0.221 & \textbf{0.104}
          & \textbf{25.96} & \textbf{0.861} & \textbf{0.090} & \textbf{0.044} \\
    \hline
    \multirow{3}{*}{Fortress}
    & 3DGS\pub{SIGGRAPH23} & 15.70 & 0.325 & 0.379 & 0.203
          & 18.80 & 0.437  & 0.367 & 0.153
          & 23.74 & 0.613 & 0.229 & 0.084 \\
    & SCG\pub{NeurIPS24} & 21.34 & 0.558 & 0.294 & 0.114
          & 25.00 & 0.823 & 0.122 & 0.056
          & 27.48 & 0.867 & 0.094 & 0.040 \\
    & HBSplat(Ours) & \textbf{22.17} & \textbf{0.646} & \textbf{0.274} & \textbf{0.097}
          & \textbf{25.68} & \textbf{0.836} & \textbf{0.117} & \textbf{0.051}
          & \textbf{27.60} & \textbf{0.871} & \textbf{0.093} & \textbf{0.039} \\
    \hline
    \multirow{3}{*}{Horns}
    & 3DGS\pub{SIGGRAPH23} & 14.35 & 0.291 & 0.511 & 0.250
          & 15.65 & 0.391  & 0.463 & 0.214
          & 19.07 & 0.528 & 0.359 & 0.145 \\
    & SCG\pub{NeurIPS24} & 16.84 & 0.554 & 0.372 & 0.176
          & 19.86 & 0.737 & 0.221 & 0.109
          & 22.71 & 0.821 & 0.158 & 0.077 \\
    & HBSplat(Ours) & \textbf{17.76} & \textbf{0.624} & \textbf{0.319} & \textbf{0.152}
          & \textbf{20.33} & \textbf{0.746} & \textbf{0.219} & \textbf{0.105}
          & \textbf{23.38} & \textbf{0.832} & \textbf{0.145} & \textbf{0.070} \\
    \hline
    \multirow{3}{*}{Leaves}
    & 3DGS\pub{SIGGRAPH23} & 13.18 & 0.249 & 0.452 & 0.266
          & 14.94 & 0.342  & 0.410 & 0.220
          & 18.20 & 0.528 & 0.300 & 0.145 \\
    & SCG\pub{NeurIPS24} & 15.30 & 0.419 & 0.454 & 0.216
          & 17.97 & 0.651 & 0.240 & 0.131
          & 19.46 & 0.720 & 0.203 & 0.106 \\
    & HBSplat(Ours) & \textbf{15.86} & \textbf{0.482} & \textbf{0.396} & \textbf{0.192}
          & \textbf{18.14} & \textbf{0.659} & \textbf{0.237} & \textbf{0.128}
          & \textbf{19.70} & \textbf{0.734} & \textbf{0.194} & \textbf{0.102} \\
    \hline
    \multirow{3}{*}{Orchids}
    & 3DGS\pub{SIGGRAPH23} & 14.88 & 0.214 & 0.492 & 0.242
          & 15.95 & 0.313  & 0.460 & 0.213
          & 20.21 & 0.472 & 0.340 & 0.132 \\
    & SCG\pub{NeurIPS24} & 14.40 & 0.384 & 0.363 & 0.218
          & 16.43 & 0.530 & 0.248 & 0.157
          & 17.93 & 0.619 & 0.203 & 0.126 \\
    & HBSplat(Ours) & \textbf{14.63} & \textbf{0.414} & \textbf{0.328} & \textbf{0.205}
          & \textbf{16.67} & \textbf{0.543} & \textbf{0.234} & \textbf{0.150}
          & \textbf{17.83} & \textbf{0.617} & \textbf{0.202} & \textbf{0.127} \\
    \hline
    \multirow{3}{*}{Room}
    & 3DGS\pub{SIGGRAPH23} & 12.09 & 0.326 & 0.581 & 0.308
          & 13.51 & 0.451  & 0.546 & 0.262
          & 16.23 & 0.451 & 0.508 & 0.207 \\
    & SCG\pub{NeurIPS24} & 19.87 & 0.793 & 0.228 & 0.104
          & 22.12 & 0.858 & 0.150 & 0.071
          & 26.20 & 0.910 & 0.107 & 0.045 \\
    & HBSplat(Ours) & \textbf{20.75} & \textbf{0.823} & \textbf{0.190} & \textbf{0.089}
          & \textbf{22.53} & \textbf{0.868} & \textbf{0.141} & \textbf{0.066}
          & \textbf{26.64} & \textbf{0.915} & \textbf{0.103} & \textbf{0.043} \\
    \hline
    \multirow{3}{*}{Trex}
    & 3DGS\pub{SIGGRAPH23} & 11.20 & 0.225 & 0.552 & 0.332
          & 12.35 & 0.344  & 0.518 & 0.290
          & 14.67 & 0.436 & 0.436 & 0.223 \\
    & SCG\pub{NeurIPS24} & 18.40 & 0.665 & 0.291 & 0.138
          & 20.62 & 0.774 & 0.193 & 0.093
          & 23.24 & 0.851 & 0.132 & 0.063 \\
    & HBSplat(Ours) & \textbf{19.35} & \textbf{0.702} & \textbf{0.236} & \textbf{0.115}
          & \textbf{21.27} & \textbf{0.787} & \textbf{0.171} & \textbf{0.084}
          & \textbf{24.01} & \textbf{0.864} & \textbf{0.120} & \textbf{0.057} \\
    \hline
    \multirow{3}{*}{Mean}
    & 3DGS\pub{SIGGRAPH23} & 13.95 & 0.288 & 0.489 & 0.255
          & 15.42 & 0.383  & 0.463 & 0.218
          & 20.41 & 0.557 & 0.322 & 0.124 \\
    & SCG\pub{NeurIPS24} & \cellcolor{myorange}18.17 & \cellcolor{myorange}0.576 & \cellcolor{myorange}0.328 & \cellcolor{myorange}0.148
          & \cellcolor{myorange}20.73 & \cellcolor{myorange}0.725 & \cellcolor{myorange}0.196 & \cellcolor{myorange}0.095
          & \cellcolor{myorange}23.38 & \cellcolor{myorange}0.809 & \cellcolor{myorange}0.139 & \cellcolor{myorange}0.065 \\

    & HBSplat(Ours) & \cellcolor{myred}18.84 & \cellcolor{myred}0.622 & \cellcolor{myred}0.280 & \cellcolor{myred}0.131
          & \cellcolor{myred}21.13 & \cellcolor{myred}0.735 & \cellcolor{myred}0.189 & \cellcolor{myred}0.090
          & \cellcolor{myred}23.71 & \cellcolor{myred}0.816 & \cellcolor{myred}0.133 & \cellcolor{myred}0.062 \\
    \hline
    \end{tabular}
    \label{tab:ll-t-235}
\end{table*}

\subsection{Comparison}
\subsubsection{LLFF \& IBRNet}
Quantitative results on the LLFF and IBRNet datasets under extremely sparse settings (3 training views) are summarized in Table \ref{tab:ll}. HBSplat achieves state-of-the-art performance, reaching a PSNR of 21.13 dB and LPIPS of 0.189 on LLFF, and 22.19 dB PSNR with 0.249 LPIPS on IBRNet, outperforming all baseline methods across metrics. Detailed per-scene results on LLFF (\ref{tab:ll-t-235}) further validate the advantage of HBSplat in forward-facing scene reconstruction.

Qualitative results are shown in Figures \ref{ll1} and \ref{ib-p}. The GT depth map is obtained from monocular depth estimator. In the Fortress and Horns scenes of LLFF dataset, SID, MCGS, and SCGaussian exhibit various artifacts and detail loss. Although SID preserves depth details better, this comes at the cost of generating more spatial points and longer training time. In the Signboard and Table scenes of IBRNet dataset, methods such as FSGS, DNGaussian, and SCGaussian also suffer from artifacts and a lack of fine details. SCGaussian is particularly prone to artifacts due to insufficient outlier filtering. Moreover, SID, MCGS, FSGS, and DNGaussian all show a common limitation: the loss of fine details in close-range areas and the introduction of blur in distant regions. 

In contrast, HBSplat produces renderings that closely match the GT across all forward-facing scenes, with enhanced edge sharpness, improved texture preservation, and minimal artifacts. These results demonstrate HBSplat's robustness and superiority in challenging sparse-view reconstruction of forward-facing scenes.

\subsubsection{Blender}
Quantitative comparisons on the Blender dataset with 8 training views are presented in Table \ref{tab:bl-t},  where HBSplat achieves competitive results across all metrics.

\begin{figure}
\centering

\includegraphics[width=3.4in]{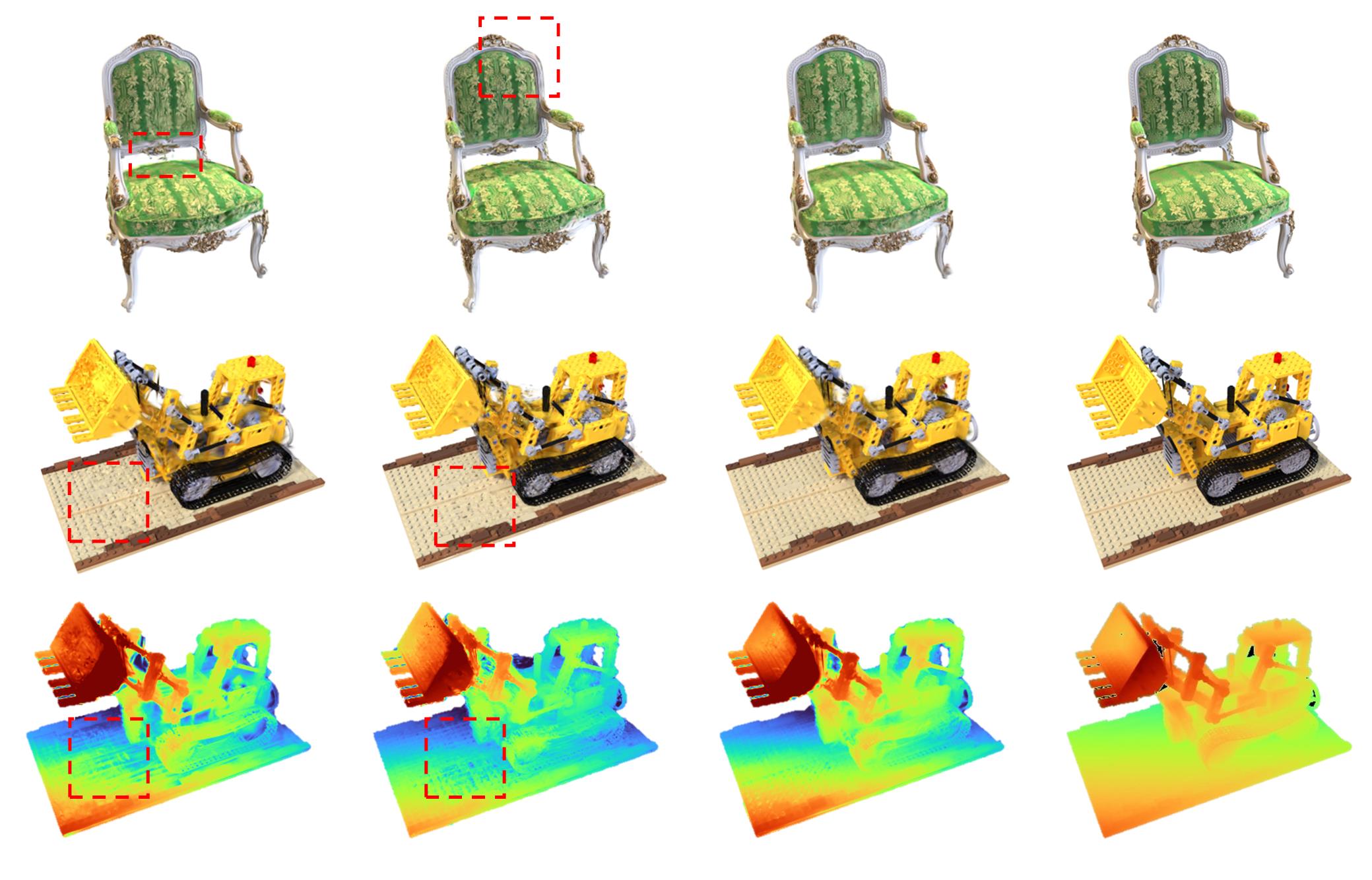}
\begin{center}
\vspace{-1mm} 
\footnotesize  
 MCGS\hspace{10mm}
 DNGaussian\hspace{7mm}
 HBSplat(Ours)\hspace{11mm}
 GT
\end{center}

\includegraphics[width=3.4in]{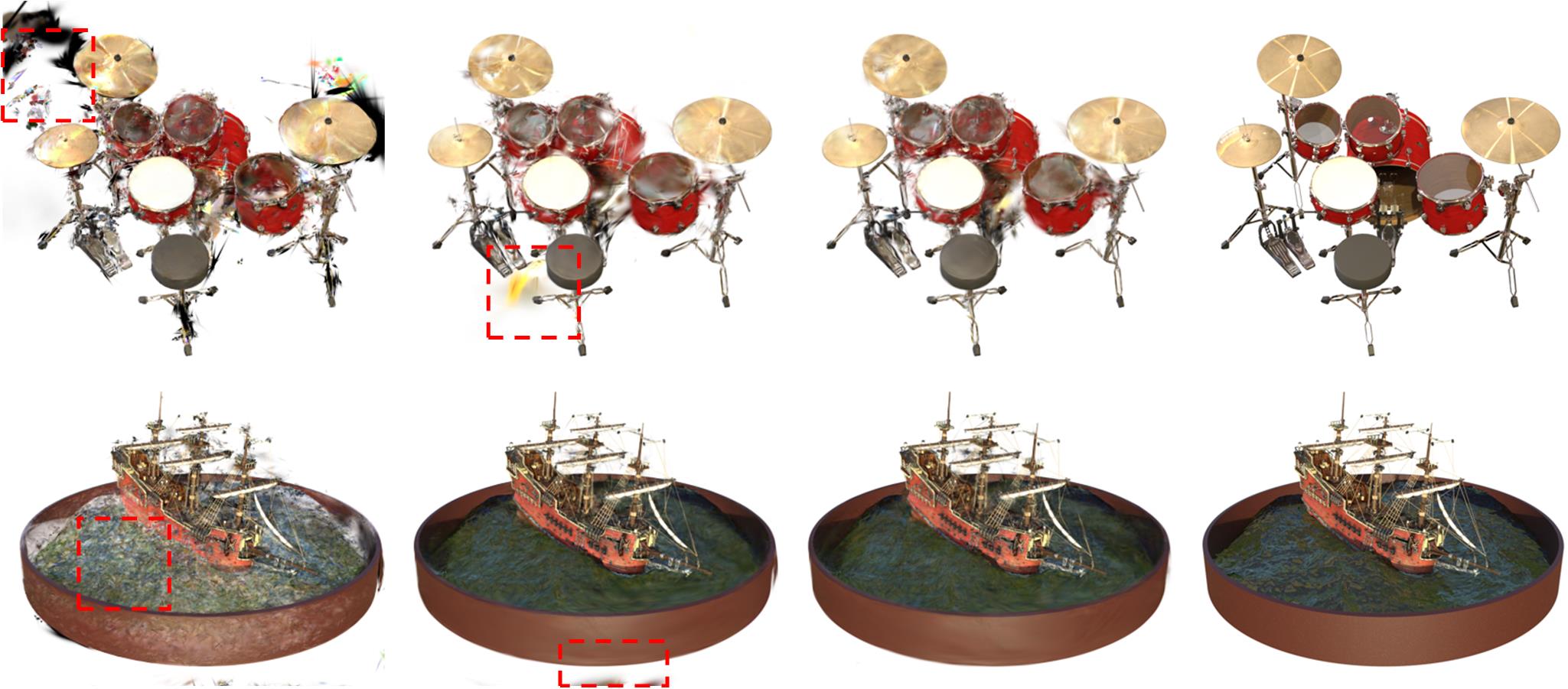}
\begin{center}
\vspace{-1mm} 
\footnotesize  
3DGS\hspace{10mm}
SCGaussian\hspace{7mm}
HBSplat(Ours)\hspace{11mm}
GT
\end{center}
\vspace{-3mm}
\caption{Qualitative comparison on Blender dataset with 8 training views. HBSplat method excels both in geometry and rendering qualities.}
\label{bl-f}
\end{figure}

\begin{table}

\centering
    \caption{Quantitative comparisons on the Blender (1/2 resolution) datasets with 8 training views. Top-3 entries are highlighted: red (1st), orange (2nd), yellow (3rd).}
    \vspace{-1mm}
    \renewcommand\arraystretch{1.3}
    \begin{tabular}{l|>{\centering\arraybackslash}p{0.86cm}|>{\centering\arraybackslash}p{0.86cm}|>{\centering\arraybackslash}p{0.86cm}|>{\centering\arraybackslash}p{0.86cm}}
    
    \hline
    \multicolumn{1}{c|}{\multirow{2}{*}{Method}} & \multicolumn{4}{c}{Blender} \\
    \cline{2-5}
    & PSNR$\uparrow$ & SSIM$\uparrow$ & LPIPS$\downarrow$ & AVG$\downarrow$ \\
    \hline
    FreeNeRF\pub{CVPR23}\cite{ref11}        & \cellcolor{myred}24.26 & 0.883 & 0.098 & \cellcolor{myyellow}0.050 \\
    SparseNeRF\pub{ICCV23}\cite{ref9}       & 22.41 & 0.861 & 0.119 & 0.063 \\ 
    \hline
    3DGS\pub{SIGGRAPH23}\cite{ref2}         & 22.85 & 0.836 & 0.141 & 0.066 \\
    FSGS\pub{ECCV24}\cite{ref20}        & 16.35 & 0.601 & 0.369 & 0.175 \\
    SID\pub{ICASSP25}\cite{ref39}         & 12.65 & 0.729 & 0.321 & 0.208 \\
    DNGaussian\pub{CVPR24}\cite{ref22}  & 22.79 & 0.870 & 0.106 & 0.058 \\
    MCGS\pub{TPAMI25}\cite{ref32}        & \cellcolor{myyellow}24.06 & \cellcolor{myorange}0.887 & \cellcolor{myorange}0.089 & \cellcolor{myorange}0.048 \\
    SCGaussian\pub{NeurIPS24}\cite{ref36}  & 23.33 & \cellcolor{myyellow}0.883 & \cellcolor{myyellow}0.097 & 0.053 \\
    \hline
    HBSplat(Ours)              & \cellcolor{myorange}24.18 & \cellcolor{myred}0.894 & \cellcolor{myred}0.086 & \cellcolor{myred}0.047  \\
    \hline
    \end{tabular}
    \label{tab:bl-t}
    \vspace{-2mm}
\end{table}

Qualitative results are shown in Figure \ref{bl-f}. In the Chair object, MCGS produces noticeable floaters, while DNGaussian suffers from blurred details. For the Lego object, both MCGS and DNGaussian fail to reconstruct fine structures, leading to a loss of detail. In the Drums and Ship objects, 3DGS and SCGaussian exhibit floters, with 3DGS showing the most severe floters as well as a clear color discrepancy from the GT. In contrast, HBSplat produces superior reconstruction results and more accurate depth maps, with sharper geometry and better-preserved textures.

\begin{figure}
\centering
\includegraphics[width=3.4in]{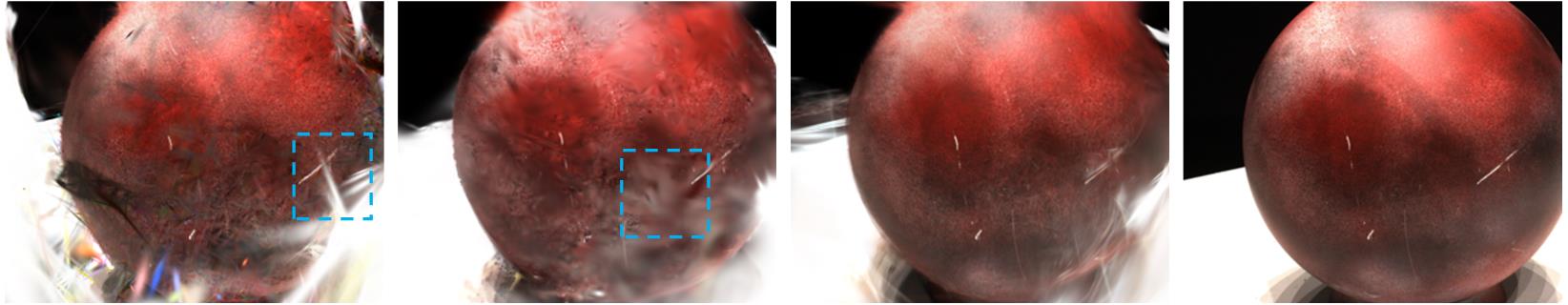}

\begin{center}
\vspace{-1mm} 
\footnotesize
 3DGS \hspace{8mm}
 DNGaussian \hspace{6mm}
 HBSplat(Ours) \hspace{10mm}
 GT
\end{center}

\includegraphics[width=3.4in]{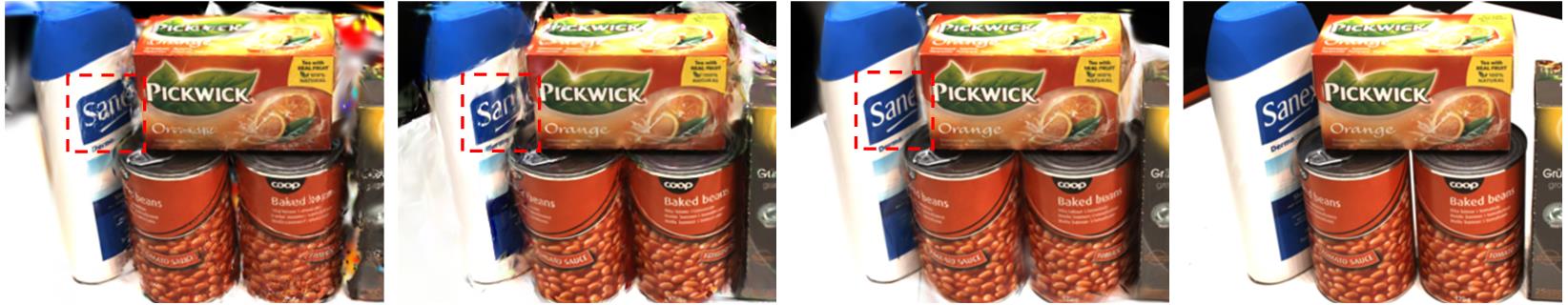}
\begin{center}
\vspace{-1mm} 
\footnotesize  
 FSGS \hspace{12mm}
 SID \hspace{11mm}
 HBSplat(Ours) \hspace{10mm}
 GT
\end{center}

\includegraphics[width=3.4in]{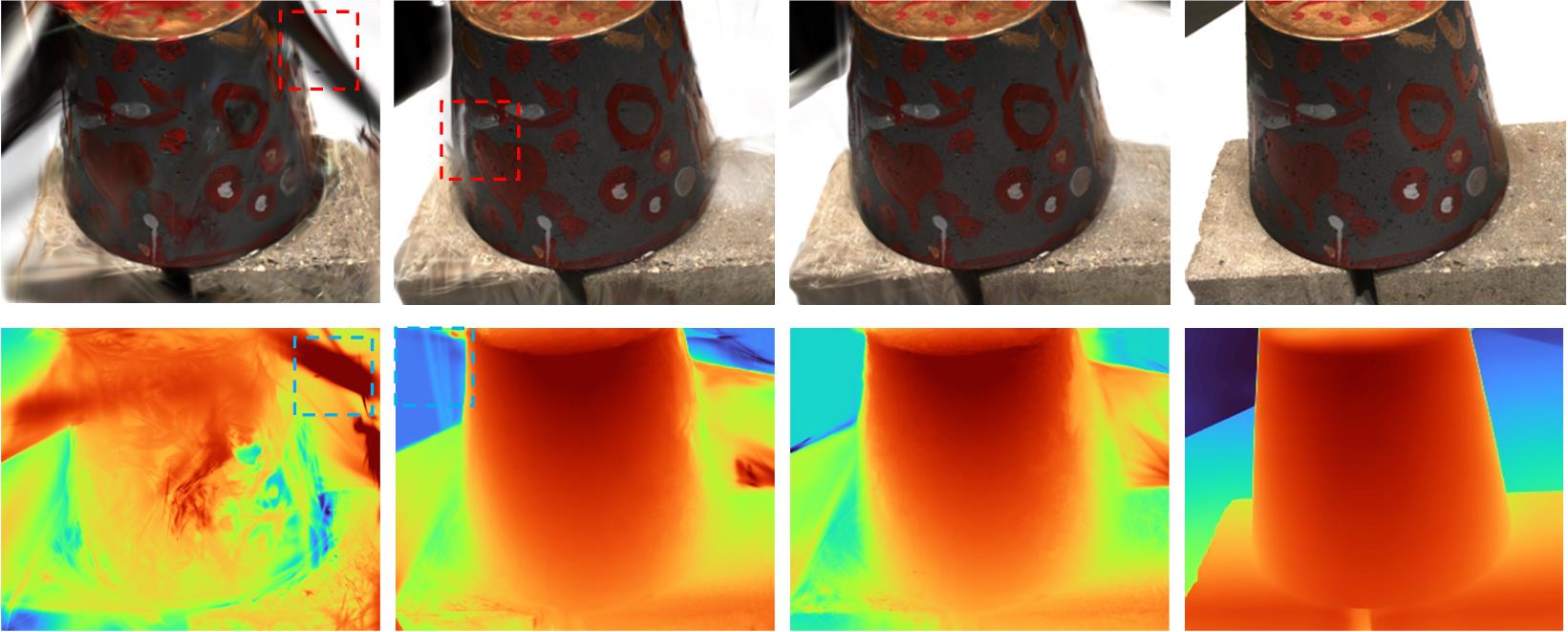}
\begin{center}
\vspace{-1mm} 
\footnotesize  
 MCGS\hspace{8mm}
 SCGaussian \hspace{6mm}
 HBSplat(Ours) \hspace{10mm}
 GT
\end{center}
\vspace{-3mm}
\caption{Qualitative comparison on DTU dataset with 3 training views.}
\label{dt-f}
\end{figure}

\subsubsection{DTU}
Quantitative comparisons on the DTU dataset are summarized in Table \ref{tab:dtu-t}, where HBSplat achieves the highest scores across all metrics, followed by SCGaussian and MCGS.

Figure \ref{dt-f} presents the rendering results. Both 3DGS and DNGaussian produce a considerable number of floaters. FSGS and SID fail to reconstruct textual details and suffer from overall blurriness. While MCGS and SCGaussian also exhibit floaters, MCGS shows the most severe cases, which further lead to inaccurate depth estimations. Benefiting from the Outlier Filtering Mechanism and virtual view constraints, HBSplat significantly reduces artifacts and improves texture fidelity, particularly around text and fine structures.

\begin{table}
\centering
    \caption{Quantitative comparisons on the DTU (1/4 resolution) datasets with 3 training views. Top-3 entries are highlighted: red (1st), orange (2nd), yellow (3rd).}
    
    \renewcommand\arraystretch{1.3}
    \begin{tabular}{l|>{\centering\arraybackslash}p{0.86cm}|>{\centering\arraybackslash}p{0.86cm}|>{\centering\arraybackslash}p{0.86cm}|>{\centering\arraybackslash}p{0.86cm}}
    
    \hline
    \multicolumn{1}{c|}{\multirow{2}{*}{Method}} & \multicolumn{4}{c}{DTU} \\
    \cline{2-5}
    & PSNR$\uparrow$ & SSIM$\uparrow$ & LPIPS$\downarrow$ & AVG$\downarrow$ \\ 
    \hline
    FreeNeRF\pub{CVPR23}\cite{ref11}        & 19.92 & 0.787 & 0.182 & 0.098 \\
    SparseNeRF\pub{ICCV23}\cite{ref9}       & 19.55 & 0.769 & 0.201 & 0.102 \\
    \hline
    3DGS\pub{SIGGRAPH23}\cite{ref2}             & 12.76 & 0.595 & 0.376 & 0.233 \\
    FSGS\pub{ECCV24}\cite{ref20}            & 17.41 & 0.728 & 0.247 & 0.132 \\
    SID\pub{ICASSP25}\cite{ref20}             & 16.35 & 0.654 & 0.334 & 0.165 \\
    DNGaussian\pub{CVPR24}\cite{ref22}      & 18.46 & 0.807 & 0.168 & 0.101 \\
    MCGS\pub{TPAMI25}\cite{ref32}            & \cellcolor{myyellow}19.02 & \cellcolor{myyellow}0.810 & \cellcolor{myyellow}0.154 & \cellcolor{myyellow}0.094 \\
    SCGaussian\pub{NeurIPS24}\cite{ref36}      & \cellcolor{myorange}19.11 & \cellcolor{myorange}0.857 & \cellcolor{myorange}0.123 & \cellcolor{myorange}0.082 \\ 
    \hline
    HBSplat(Ours)                  & \cellcolor{myred}20.22 & \cellcolor{myred}0.872 & \cellcolor{myred}0.110 & \cellcolor{myred}0.072  \\ 
    \hline
    \end{tabular}
    \label{tab:dtu-t}
    \vspace{-1mm}
\end{table}

\subsubsection{Tanks\&Temples}
The Tanks\&Temples dataset is employed to evaluate the performance of HBSplat in large-scale 360° unbounded scenes. Quantitative results are presented in Table \ref{tab:tt-t}. Note that evaluations on this dataset utilize only the Hybrid-Loss Depth Estimation component of our framework. DNGaussian and MCGS are excluded from comparison as they lack configurations suitable for 360° unbounded scenes.

\begin{figure*}
\centering
\includegraphics[width=7in]{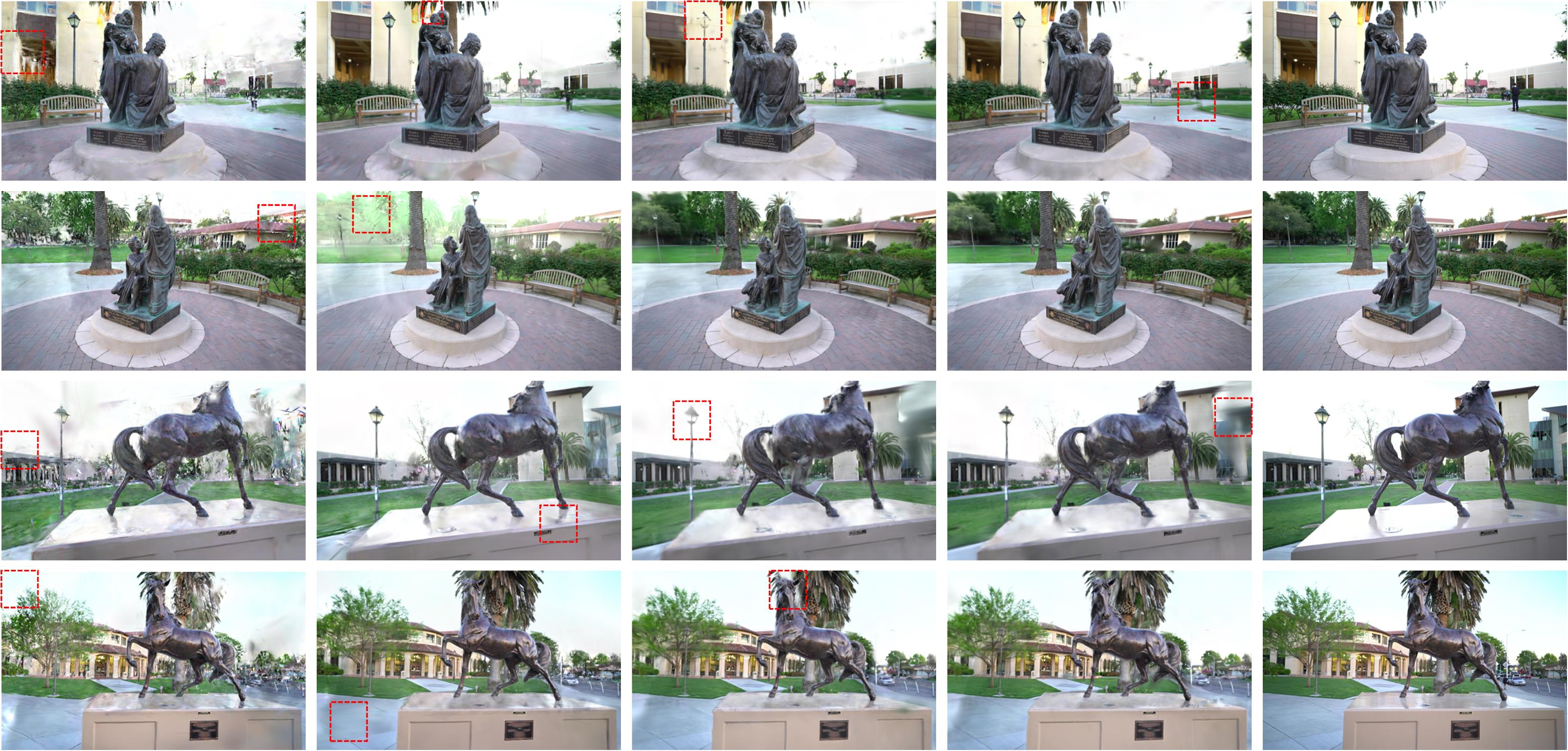}

\begin{center}
\footnotesize  
 3DGS\hspace{30mm}
 SID\hspace{27mm}
 SCGaussian\hspace{21mm}
 HBSplat(Ours)\hspace{23mm}
 GT
\end{center}
\vspace{-3mm} 
\caption{Qualitative comparison on Tanks\&Temples dataset with 24 training views. HBSplat method excels both in geometry and rendering qualities.}
\label{tt-f}
\end{figure*}

Qualitative results are shown in Figure \ref{tt-f}. As SID is an enhanced version of FSGS, only SID is presented for clarity. In the Family and Horse scenes, both vanilla 3DGS and SID exhibit significant artifacts, with the latter also showing noticeable color rendering errors. SCGaussian suffers from partial object omission. In contrast, HBSplat produces more complete and geometrically stable reconstructions with reduced artifacts, demonstrating its robustness in challenging large-scale unbounded scenarios.

\begin{table}[t]
\vspace{-3mm}
\centering
    \caption{Quantitative comparisons on the Tanks\&Temples (1/4 resolution) datasets with 24 training views. Top-3 entries are highlighted: red (1st), orange (2nd), yellow (3rd).}
    
    \renewcommand\arraystretch{1.3}
    \begin{tabular}{l|>{\centering\arraybackslash}p{0.86cm}|>{\centering\arraybackslash}p{0.86cm}|>{\centering\arraybackslash}p{0.86cm}|>{\centering\arraybackslash}p{0.86cm}}
    
    \hline
    \multicolumn{1}{c|}{\multirow{2}{*}{Method}} & \multicolumn{4}{c}{Tanks\&Temples} \\
    \cline{2-5}
    & PSNR$\uparrow$ & SSIM$\uparrow$ & LPIPS$\downarrow$ & AVG$\downarrow$ \\ 
    \hline
    3DGS\pub{SIGGRAPH23}\cite{ref2}         & 18.77 & 0.706 & 0.267 & 0.124 \\
    FSGS\pub{ECCV24}\cite{ref20}        & \cellcolor{myyellow}19.39 & 0.720 & \cellcolor{myyellow}0.273 & \cellcolor{myyellow}0.118 \\
    SID\pub{ICASSP25}\cite{ref20}         & \cellcolor{myorange}19.51 & \cellcolor{myorange}0.723 & \cellcolor{myorange}0.273 & \cellcolor{myorange}0.117 \\
    SCGaussian\pub{NeurIPS24}\cite{ref36}  & 18.96 & \cellcolor{myyellow}0.722 & 0.298 & 0.125 \\
    \hline
    HBSplat(Ours)              & \cellcolor{myred}20.00 & \cellcolor{myred}0.749 & \cellcolor{myred}0.268 & \cellcolor{myred}0.110  \\ 
    \hline
    \end{tabular}
    \label{tab:tt-t}
    \vspace{-3mm}
\end{table}

\subsubsection{Efficiency}
Efficiency evaluation of different methods on the LLFF dataset in Table \ref{tab:effi}. The proposed HBSplat method matches the efficiency of other advanced approaches, achieving an inference speed of 250 FPS, which is comparable to methods such as SCGaussian and MCGaussian, while maintaining a low average training time of 2.5 minutes. For GPU memory, HBSplat requires 3 GB for training and an additional 5 GB for the Virtual View Synthesis (VVS) component to generate 100 virtual views, outperforming methods like FreeNeRF (4×48 GB) and SparseNeRF (32 GB). Compared to approaches such as 3DGS (7.5 minutes, 2 GB) and FSGS (17 minutes, 3 GB), HBSplat offers a substantial reduction in training time while preserving real-time inference capabilities, making it a highly efficient solution for practical 3D reconstruction tasks.

\begin{table}
    \centering
    \caption{Efficiency comparison of different methods on LLFF dataset.}
    \renewcommand\arraystretch{1.3}
    \begin{tabular}{l|c|c|c}
    \hline
    \multicolumn{1}{c|}{Method} & \makecell{\rule{0pt}{2.5ex}Inference\\(FPS)} & \makecell{\rule{0pt}{2.5ex}Average \\Training \\Time} & GPU Mem  \\ 
    \hline
    FreeNeRF\pub{CVPR23}\cite{ref11} & $9 \times 10^{-2}$ & 2.3 h & 4$\times$48 GB \\
    SparseNeRF\pub{ICCV23}\cite{ref9} & $9 \times 10^{-2}$ & 1.5 h & 32 GB \\
    \hline
    3DGS\pub{SIGGRAPH23}\cite{ref2}             & 400 & 7.5 min & 2 GB\\
    FSGS\pub{ECCV24}\cite{ref20}            & 300 & 17 min & 3 GB \\
    SID\pub{ICASSP25}\cite{ref20}             & 300 & 28 min & 3.5 GB \\
    DNGaussian\pub{CVPR24}\cite{ref22}      & 500 & 3.5 min & 2 GB \\
    MCGS\pub{TPAMI25}\cite{ref32}            & 250 & 2.5 min & 2 GB \\
    SCGaussian\pub{NeurIPS24}\cite{ref36}      & 250 & 1.5 min & 3 GB \\ 
    \hline
    HBSplat(Ours)                        & 250 & 2.5 min & \makecell[c]{\rule{0pt}{2.5ex}3 GB
    \\5 GB\ (VVS)} \\ 
    \hline
    \end{tabular}
    \label{tab:effi}
    \vspace{-3mm}
\end{table}

\subsection{Ablation Study}
Ablation study is conducted on the LLFF dataset under a 3-view setting to evaluate the individual contributions of the core components in HBSplat: Hybrid-Loss Depth Estimation (HLDE), Bidirectional Warping Virtual View Synthesis (VVS), and Occlusion-Aware Reconstruction (OAR). 

\textbf{Quantitative Component Analysis:}
Quantitative results are summarized in Table \ref{tab:ablation}.  
The full HBSplat model (HLDE+VVS+OAR) yields the best overall performance, achieving a PSNR of 21.13 dB and an LPIPS of 0.189. 
Among the components, VVS contributes the highest PSNR due to the strong photometric constraints imposed by virtual views. Although HLDE yields a lower PSNR than VVS, it infers more accurate geometry, resulting in the lowest LPIPS. Notably, each individual component of HBSplat outperforms all other baselines. Furthermore, the performance gain from combining components is non-linear, integrating HLDE with VVS alone is sufficient to achieve substantial improvement.

\begin{table}
\vspace{-3mm}
\centering
\caption{Ablation study on the components of HBSplat.}
\label{tab:performance}
\renewcommand\arraystretch{1.2}
\begin{tabular}{>{\raggedright\arraybackslash}p{1.1cm}|>{\centering\arraybackslash}p{0.5cm}|>{\centering\arraybackslash}p{0.4cm}|>{\centering\arraybackslash}p{0.4cm}|>{\centering\arraybackslash}p{0.7cm}|>{\centering\arraybackslash}p{0.7cm}|>{\centering\arraybackslash}p{0.7cm}|>{\centering\arraybackslash}p{0.7cm}
}
\hline
\multicolumn{1}{c|}{Method} & \text{\hspace{-1mm}HLDE} & \text{\hspace{-0.5mm}VVS} & \text{\hspace{-0.5mm}OAR} & \text{\hspace{-0.5mm}PSNR$\uparrow$} & \text{\hspace{-0.5mm}SSIM$\uparrow$} & \text{\hspace{-0.5mm}LPIPS$\downarrow$} & AVG$\downarrow$\\ 
\hline
\text{\hspace{-1mm}3DGS}    &  \redcross &  \redcross &  \redcross & 15.42 & 0.383 & 0.463 & 0.221\\ 
\text{\hspace{-1mm}SCGaussian} &  \redcross &  \redcross &  \redcross & 20.73 & 0.725 & 0.196 & 0.100\\ 
\hline
\multirow{5}{*}{\makecell{\rule{0pt}{2.5ex}HBSplat\\(Ours)}}   
        &  \greencheck &  \redcross &  \redcross & 20.91 & 0.732 & 0.187 & 0.097\\ 
        &  \redcross &  \greencheck &  \redcross & 20.93 & 0.731 & 0.191 & 0.098\\ 
        &  \redcross &  \redcross &  \greencheck & 20.85 & 0.727 & 0.191 & 0.099\\ 
        &  \greencheck &  \greencheck &  \redcross &  21.11 & 0.734 & 0.191 & 0.096\\
        &  \greencheck &  \greencheck &  \greencheck & 21.13 & 0.735 & 0.189 & 0.096\\
\hline

\end{tabular}
\label{tab:ablation}
\vspace{-3mm}
\end{table}

\begin{figure}
\centering

\includegraphics[width=3.4in]{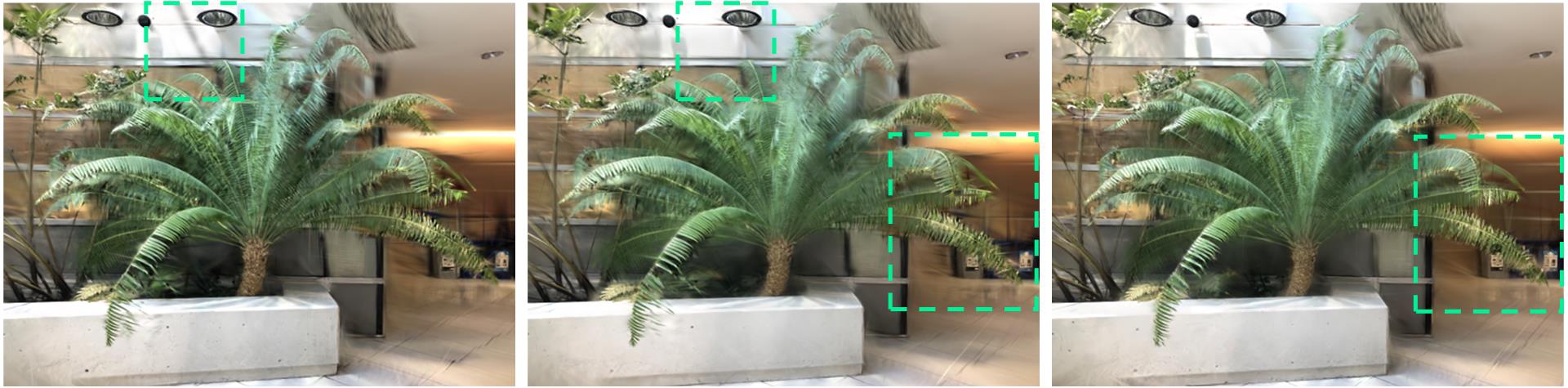}
\begin{center}
\vspace{-2mm} 
\footnotesize
\qquad
\hspace{2mm}
+HLDE \hspace{13mm}
+HLDE+VVS\hspace{9mm}
+HLDE+VVS+OAR
\end{center}

\includegraphics[width=3.4in]{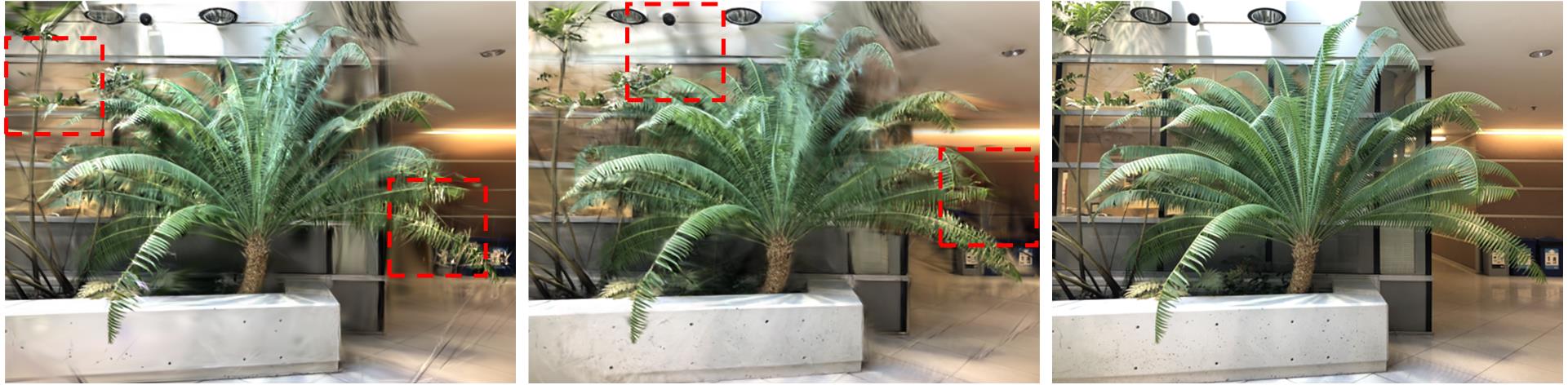}
\begin{center}
\vspace{-2mm} 
\footnotesize 
 MCGS\hspace{17mm}
 SCGaussian\hspace{19mm}
 GT
\end{center}
\vspace{-3mm}
\caption{Ablation study on component contributions. Compared to MCGS and SCGaussian, the rendering results of full HBSplat model are closest to GT.}
\label{ablation-f}
\end{figure}

\textbf{Qualitative Component Analysis:}
Figure \ref{ablation-f} provides a qualitative comparison on the Fern scene. The following observations are made in the boxed regions:
+HLDE: Produces more robust depth estimates, effectively resolving structural distortions present in 3DGS.
+VVS: Introduces virtual view constraints, significantly reducing artifacts and improving consistency.
+OAR: Generates more plausible background content and alleviates occlusion-related artifacts. 

In the second row of the comparison, MCGS reconstructs leaves as sparse, needle-like structures and suffers from blurred edges. SCGaussian exhibits black artifacts and fails to reconstruct the background completely. In contrast, HBSplat produces renderings that are visually closer to the GT, demonstrating the  effectiveness of the proposed components.

\begin{figure}
\centering
\includegraphics[width=3.4in]{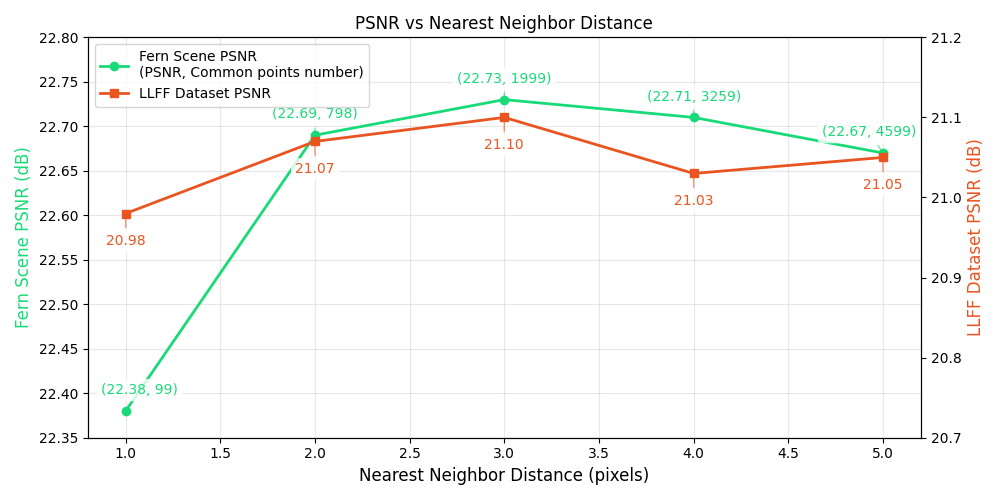}

\caption{PSNR vs. $d_{NN}$. Both the number of common points and PSNR initially increase with $d_{NN}$ but decline after the $d_{NN}$ exceeds 3.}
\vspace{-3mm}
\label{nnd}
\end{figure}

\textbf{PPC with different nearest neighbor distance ($d_{NN}$):}
Figure \ref{nnd} shows the line graph of PSNR values over different $d_{NN}$. It can be observed that as $d_{NN}$ increases, both the number of common points and the PSNR value gradually rise. However, when the $d_{NN}$ exceeds 3, the PSNR for both the Fern scene and the LLFF dataset begins to decline. The point (22.38, 99) represents the PSNR value and the number of common points for the Fern scene, respectively. Notably, if the number of common points is less than 100 following a secondary filtering step, the PPC operation is skipped to avoid redundant processing.

\section{Conclusion}
This paper presents HBSplat, an effective framework for sparse-view NVS based on 3DGS. The core of our approach includes three innovations: a Hybrid-Loss Depth Estimation module that enforces multi-view consistency through reprojection, point propagation, and TV smoothness constraints; a Bidirectional Warping Virtual View Synthesis method that enhances the coverage of unobserved regions, thereby mitigating overfitting to the limited input views; and an Occlusion-Aware Reconstruction component that improves background rendering through depth-difference priors. Extensive experiments across standard benchmarks demonstrate that HBSplat achieves state-of-the-art performance under extreme sparsity. We believe our work offers a valuable step toward practical and high-quality 3D reconstruction from very few images.

\textbf{Limitations and Future Work:}
HBSplat has several limitations. It relies on COLMAP for pose estimation and uses fixed propagation thresholds, potentially limiting adaptability. The occlusion module, optimized for foreground objects, may struggle with large homogeneous regions such as sky. Additionally, it only supports static scenes. Future work will explore self-supervised pose estimation, adaptive thresholds, and dynamic scene modeling to address these constraints.

\bibliographystyle{IEEEtran}
\bibliography{IEEEabrv,references}

\end{document}